\documentclass[11pt]{article}

\usepackage[margin=1in]{geometry}
\usepackage[utf8]{inputenc}
\usepackage[T1]{fontenc}
\usepackage{lmodern}
\usepackage{xcolor}     
\usepackage{pifont}   
\usepackage{multirow}
\usepackage{booktabs} 
\usepackage{fancyvrb}
\usepackage{underscore}
\usepackage{siunitx} 
\usepackage{comment}
\usepackage{xcolor}
\usepackage{comment}
\definecolor{tiiPurple}{RGB}{122, 0, 255}

\usepackage{pgf-pie}
\usepackage{xcolor}
\usepackage{tcolorbox}
\usepackage{pgfplots}
\usepackage{threeparttable} 
\usepackage{float}       
\usepackage{minted}      
\usepackage{algorithm}   
\usepackage{algpseudocode} 

\pgfplotsset{compat=1.18}

\usepackage{sidecap}

\definecolor{mathblue}{RGB}{70, 130, 180}
\definecolor{codeorange}{RGB}{255, 140, 0}
\definecolor{othergreen}{RGB}{60, 179, 113}
\definecolor{multiturnpurple}{RGB}{147, 112, 219}
\definecolor{ifRed}{RGB}{220, 20, 60}
\definecolor{sciencebrown}{RGB}{139, 90, 43}
\definecolor{logicpink}{RGB}{255, 182, 193}

\usepackage{hyperref}
\usepackage{subcaption} 
\hypersetup{
    colorlinks=true,
    linkcolor=tiiPurple!75!black,
    urlcolor=tiiPurple!75!black,
    citecolor=tiiPurple
}

\usepackage{graphicx}
\usepackage{tabularx}
\usepackage[useregional]{datetime2}
\DTMsetdatestyle{iso}  
\usepackage[normalem]{ulem}
\usepackage{titlesec}
\usepackage{adjustbox}
\usepackage{enumitem}
\usepackage{booktabs}
\usepackage{multirow}
\usepackage{array}
\usepackage{float}
\usepackage{caption}
\usepackage{amsfonts}       
\usepackage{amsmath, amssymb}       
\usepackage{nicefrac}       
\usepackage{hyperref}       
\usepackage{url}            
\usepackage[utf8]{inputenc} 
\usepackage[T1]{fontenc}    
\usepackage{natbib}
\usepackage{graphicx}
\usepackage{enumitem}
\usepackage{makecell}
\usepackage{tabularx}
\usepackage{pdfpages}
\usepackage{arydshln}

\usepackage{longtable}
\definecolor{bestcolor}{RGB}{220,255,220}

\definecolor{cmohamed}{RGB}{150,1,25}
\definecolor{cabdelgader}{RGB}{100,25,0}
\definecolor{ciheb}{RGB}{25,100,25}
\definecolor{cshi}{RGB}{100,25,100}
\definecolor{cpuneesh}{RGB}{0,25,100}
\definecolor{creda}{RGB}{0,100,100}
\definecolor{csuhail}{RGB}{100,100,0}
\definecolor{cslim}{RGB}{25,150,0}

\usepackage{wrapfig}

\usepackage{siunitx}   
\usepackage{booktabs}  
\usepackage{pgf-pie}    

\newcommand{\reda}[1]{\textcolor{creda}{#1}\marginpar{\footnotesize{\textcolor{creda}{Reda}}}}

\titleformat{\section}
  {\normalfont\Large\bfseries}{\thesection.}{1em}{}
\usepackage{fancyhdr}
\begin{document}
\noindent
\begin{minipage}[t]{0.49\textwidth}
    \vspace*{-4.2em}
    \includegraphics[height=1.3cm,width=2.5cm]{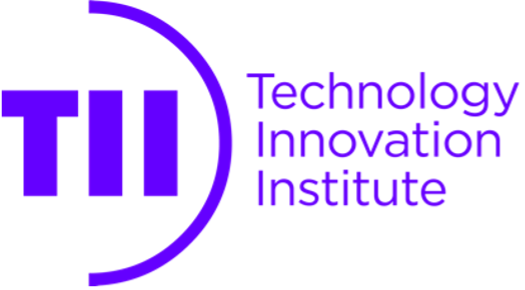}  
\end{minipage}%
\hfill
\begin{minipage}[t]{0.49\textwidth}
    \vspace*{-1.5em}
    \raggedleft
    \today
\end{minipage}
\vspace*{-0.5em}
\hrule
\vspace{1.2em}

\begin{center}
    {\Large \textbf{\textcolor{tiiPurple}{Falcon-H1R: Pushing the Reasoning Frontiers with a Hybrid Model for Efficient Test-Time Scaling}}}
\end{center}

\noindent

\begin{center}

\textbf{Falcon LLM Team}\footnote{Correspondence to \href{mailto:Falcon-LLM@tii.ae}{Falcon-LLM@tii.ae}.} \\
Iheb Chaabane \quad Puneesh Khanna \quad Suhail Mohmad \quad Slim Frikha\\
Shi Hu \quad Abdalgader Abubaker \quad Reda Alami \quad 
Mikhail Lubinets \\
Mohamed El Amine Seddik \quad Hakim Hacid
\vspace{1em}

\begin{tabular}{@{}l l@{}}
    \raisebox{-0.15\height}{\includegraphics[width=0.4cm]{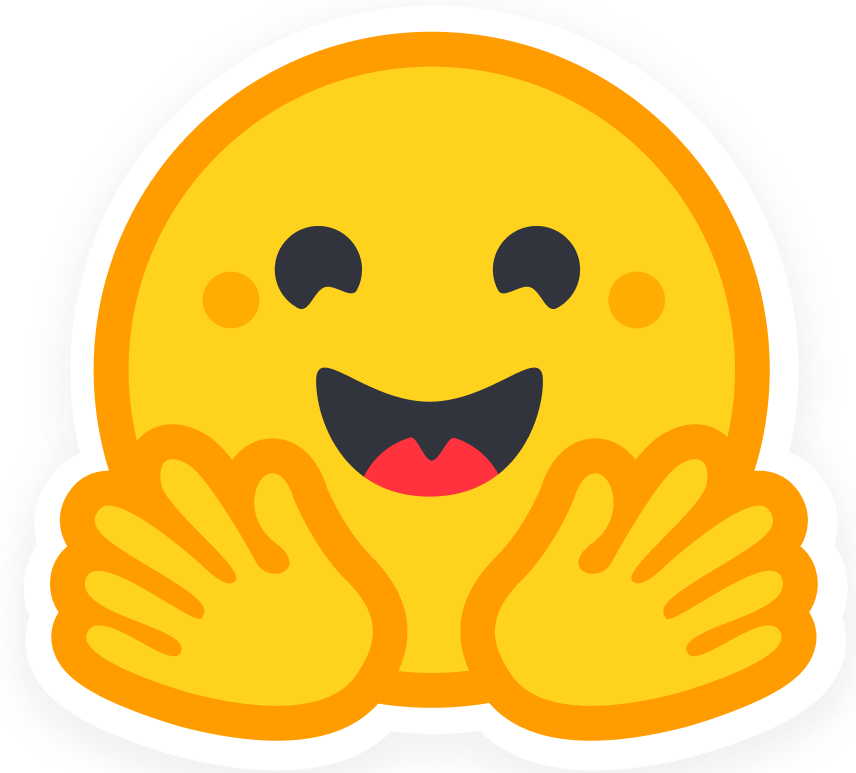}} &
    \href{https://huggingface.co/tiiuae}{https://huggingface.co/tiiuae} [\href{https://huggingface.co/collections/tiiuae/falcon-h1r}{Falcon-H1R collection}] \\
\end{tabular}

\end{center}


\pagestyle{plain}

\begin{abstract}
This work introduces Falcon-H1R, a 7B-parameter reasoning-optimized model that establishes the feasibility of achieving competitive reasoning performance with small language models (SLMs). Falcon-H1R stands out for its parameter efficiency, consistently matching or outperforming SOTA reasoning models that are $2\times$ to $7\times$ larger across a variety of reasoning-intensive benchmarks. These results underscore the importance of careful data curation and targeted training strategies (via both efficient SFT and RL scaling) in delivering significant performance gains without increasing model size. Furthermore, Falcon-H1R advances the 3D limits of reasoning efficiency by combining faster inference (through its hybrid-parallel architecture design), token efficiency, and higher accuracy. This unique blend makes Falcon-H1R-7B a practical backbone for scaling advanced reasoning systems, particularly in scenarios requiring extensive chain-of-thoughts generation and parallel test-time scaling. Leveraging the recently introduced DeepConf approach, Falcon-H1R achieves state-of-the-art test-time scaling efficiency, offering substantial improvements in both accuracy and computational cost. As a result, Falcon-H1R demonstrates that compact models, through targeted model training and architectural choices, can deliver robust and scalable reasoning performance.
\end{abstract}

\begin{figure}[ht]
\centering
\begin{minipage}{0.6\textwidth}
    \includegraphics[width=\linewidth]{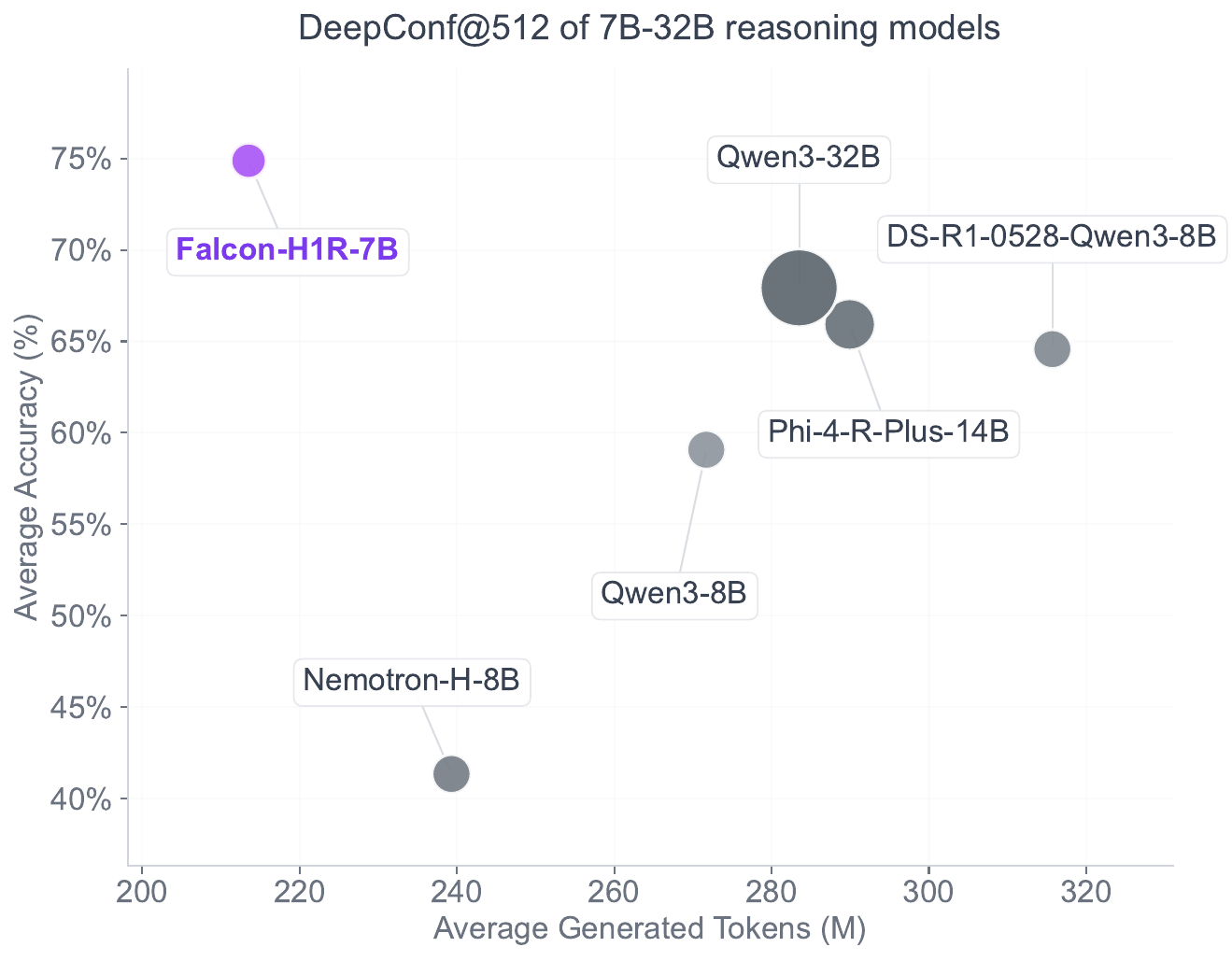}
\end{minipage}
\hspace{1em}
\begin{minipage}{0.29\textwidth}
    \captionof{figure}{\small DeepConf@512 \citep{deepconf} average results over AIME24, AIME25, AMO-Bench, and GPQA-Diamond (Detailed results in Table \ref{tab:bench}). \href{https://huggingface.co/tiiuae/Falcon-H1R-7B}{Falcon-H1R-7B} achieves exceptional performance by pushing the reasoning frontiers in 3 dimensions: higher accuracy, token efficiency and fast inference in the parallel thinking setting.}
\end{minipage}
\end{figure}

\newpage
\tableofcontents

\section{Introduction}
Large language models (LLMs) have rapidly advanced the state of complex reasoning tasks, achieving impressive results by scaling compute in two independent directions:
\begin{itemize}
    \item \textit{Scaling Training:} Improving model capabilities through comprehensive training \citep{kaplan, hoffmann} that typically involves Supervised Fine-Tuning (SFT) on high-quality instruction data \citep{guha2025openthoughts,yue2025does} followed by Reinforcement Learning (RL) (e.g., RLVR) for fine-grained alignment and performance maximization \citep{DSR1,openai2024o1,DAPO}. 
    \item \textit{Scaling Inference:} Proposing parallel thinking methods that generate and aggregate multiple solution chains, such as through self-consistency or majority voting \citep{fu2025deep, lightman2023lets, stiennon2020learning,nakano2021webgpt,uesato2022solving}.
\end{itemize}
Scaling training strategies have enabled LLMs to tackle increasingly complex reasoning tasks. However, as recent work notes \citep{wang2023selfconsistency, snell, yao2024tree, dziri}, pure pretraining progress is slowing due to extreme compute requirements and limited high-quality human data. This has motivated an emerging paradigm: Test-Time Scaling (TTS) \citep{TTSsurvey}, which allocates additional inference-time compute to unlock latent reasoning capabilities. The notable successes of reasoning models \citep{DSR1,openai2024o1,google2025geminipro} have further fueled interest in TTS, highlighting its importance for both LLM reasoning and practical utility.

TTS has shown significant gains across reasoning-intensive domains. In mathematics, sampling multiple reasoning chains and selecting consistent solutions improves accuracy \citep{wang2023selfconsistency}. In code generation, generating diverse candidates and verifying them via execution enhances functional correctness \citep{chen2024largesearchcode, dziri}. For multi-hop and scientific reasoning, search-based inference approaches like Tree-of-Thoughts boost compositional reasoning \citep{yao2024tree}, while agentic and evolutionary methods \citep{alphaevolve} extend these ideas to open-ended scientific discovery. More generally, scaling inference-time compute improves reliability and calibration through confidence-based pruning \citep{snell, fu2025deep}.

Despite these benefits, TTS incurs high inference costs, as generating and evaluating many candidate solutions per query is compute-intensive \citep{s1}. Balancing efficiency with strong baseline accuracy is therefore critical, particularly for models handling large parallel batches and long sequences. With this in mind, we build on the Falcon-H1 series \citep{falconH1}, a family of hybrid Transformer–Mamba architectures \citep{mamba,zamba2,jamba,hymba,minimax,nemotronh, aprielh1, kimilinear} optimized for high throughput and low memory usage at long sequence lengths and high batch sizes. We construct \textbf{Falcon-H1R-7B} via further SFT and RL scaling of Falcon-H1-7B, to achieve a compact model that remains competitive with 8B–32B systems while substantially reducing inference overhead.
We further incorporate a state-of-the-art TTS method that dynamically prunes weak reasoning chains during generation \citep{fu2025deep}, and demonstrate that Falcon-H1R enhances TTS efficiency by accommodating more parallel chains within the same compute budget and enabling effective early stopping. Collectively, these properties make Falcon-H1R a highly effective backbone for reasoning workloads demanding both accuracy and scalability.\\

\noindent  \textbf{Contributions}: We introduce Falcon-H1R, a 7B reasoning-optimized model leveraging a hybrid Transformer-SSM architecture for superior inference efficiency, designed to maximize the efficacy and efficiency of test-time scaling methods. The key contributions of this work are as follows:

\begin{enumerate}
    \item \textbf{Hybrid Architecture for Efficient Reasoning via TTS:} We leverage the Falcon-H1 architecture \citep[Section 2]{falconH1}, a parallel hybrid Transformer–Mamba (state-space) architecture known for its superior inference speed and memory efficiency \citep[Section 5.3]{falconH1} making it an ideal backbone for reasoning tasks that require high throughput under large batch sizes, which is typical of parallel test-time scaling techniques.

    \item \textbf{Robust Training Strategy:} We train our model via cold-start supervised fine-tuning on datasets with long reasoning traces, with capabilities further enhanced through reinforcement learning using the GRPO approach. The SFT stage leverages rigorously curated data spanning mathematics, code, and science domains, with difficulty-aware filtering to emphasize challenging problems. GRPO training builds on the SFT model and addresses distinctive challenges, including training with exceptionally large response lengths (up to 48K tokens) and balancing exploration to improve output quality. The final model achieves strong performance on popular reasoning benchmarks, including 88.1\% on AIME24, 83.1\% on AIME25, 64.9\% on HMMT25, 36.3\% on AMO-Bench, and 68.6\% on LiveCodeBenchv6 (Tables \ref{tab:eval-math}-\ref{tab:eval-code}-\ref{tab:eval-general}), competing with even larger and recent SOTA reasoning models such as GPT-OSS-20B, Qwen3-32B, Phi-4-Reasoning-Plus-14B, and DeepSeek-R1-0528-Qwen3-8B.
    
    \item \textbf{Superior Efficiency and Accuracy via TTS:} By shifting the accuracy–cost frontier, Falcon-H1R delivers state-of-the-art reasoning performance with substantially lower inference overhead, demonstrating the impact of targeted training and architectural choices in realizing the full benefits of TTS. We evaluate Falcon-H1R using the DeepConf method \citep{fu2025deep}, which dynamically filters and aggregates parallel reasoning chains based on model confidence. Our results (Table \ref{tab:bench}) demonstrate that Falcon-H1R consistently improves both accuracy and cost-efficiency in test-time scaling scenarios. On AIME25, for example, Falcon-H1R-7B attains 96.7\% accuracy while reducing token usage by 38\% relative to DeepSeek-R1-0528-Qwen3-8B. These gains stem from strong reasoning performances of the base Falcon-H1R-7B model, as well as well-calibrated confidence estimates that support aggressive early stopping without sacrificing accuracy.
\end{enumerate}

\noindent  \textbf{Organization}: The remainder of this technical report is organized to provide a clear and comprehensive overview of the steps taken to achieve our results. We begin with Section \ref{sec:SFT}, which details the supervised fine-tuning (SFT) stage, including data processing, filtering, ablation findings, key hyperparameters, and training specifics. Section \ref{sec:RL} covers the reinforcement learning (RL) stage, describing data preparation, the training framework, experimental ablations, and the overall setup. Both sections conclude with key training insights. In Section \ref{sec:eval}, we present our evaluation methodology and results across a wide range of reasoning benchmarks. Finally, Section \ref{sec:TTS} provides an in-depth discussion of the test-time scaling experiments and outcomes.

\section{Cold-start SFT Stage}\label{sec:SFT}

To further expand the base model reasoning capabilities, the current paradigm relies first on cold-start supervised fine-tuning stage. This is particularly motivated by \citep{yue2025does}, which shows that while RLVR-training does indeed improve average accuracy (e.g., at pass@1), a deeper analysis reveals that the coverage of solvable problems (measured by pass@k for large k) is narrower than or equal to the base model's, essentially all reasoning paths \textit{discovered} by the RL-training are already present in the base model's distribution. This section provides details on our SFT stage, which proved critical to performance improvements in our model, accounting for the majority of reasoning capability gains in our training pipeline.

\subsection{Data Filtering \& Processing}

We curated a diverse set of datasets spanning different domains, including mathematics, coding, STEM, instruction-following, tool calling, and general chat, from which long chain-of-thoughts have been generated. Datasets were then organized into four primary domains as depicted in Fig.\,\ref{fig:data_pie}: mathematics, coding (with an emphasis on Python and C++), science, and other, comprising the remaining domains. For coding data, we emphasized algorithmic reasoning and functional correctness. In contrast, for mathematics and science tasks, we prioritized problems with verified
ground-truth answers, resorting to an LLM-based judge only when such labels were unavailable.

We implemented rigorous data verification, including selecting mathematical and science solutions based on correctness and cross-validating factual questions using official web sources when applicable. For mathematics, we relied on both \textsc{math-verify}\footnote{\url{https://github.com/huggingface/Math-Verify}} and LLM-based verification for greater flexibility. Only correct solutions were kept, except for particularly difficult incorrect solutions or hardly verifiable problems such as theorems, which were retained to expose the models to complex reasoning traces as shown to be effective in \citep{guha2025openthoughts}.

\begin{figure}[t!]
\centering
\includegraphics[width=\textwidth]{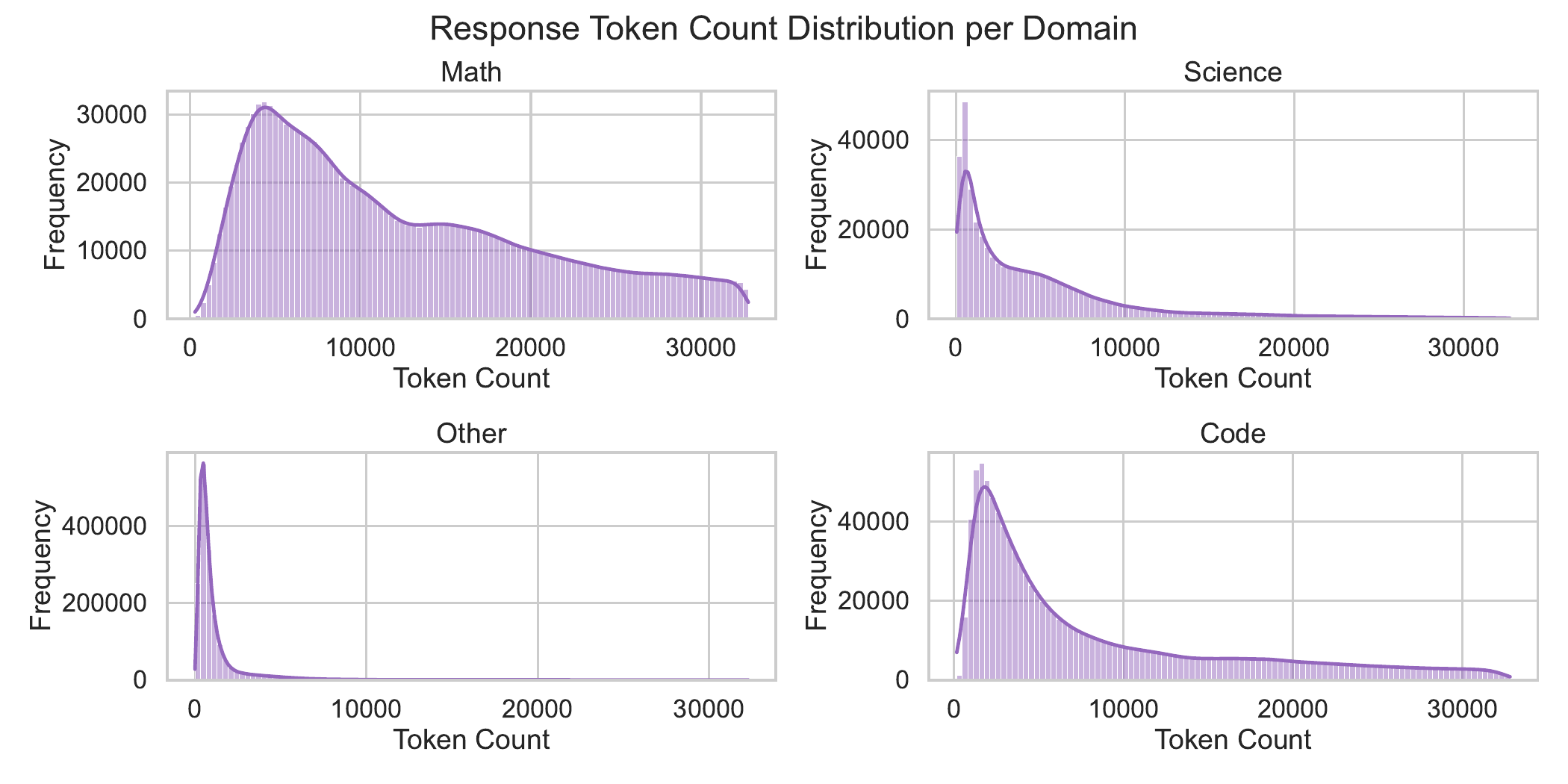}
\caption{Distribution of response token counts per domain in the SFT stage. The ``Other'' category includes IF, Chat, Safety, and Tool calling data.}
\label{fig:response_token_dist}
\end{figure}

Quality filtering was also applied using several criteria: removing instances with empty reasoning content or final answer, mathematics solutions lacking a final \text{\textbackslash boxed\{\}} answer, code with syntax errors or incorrect coding language, and tool-calling data that failed to produce reliable outputs. After filtering, the response token count distribution for each domain followed an approximately log-normal shape as shown in Fig.\,\ref{fig:response_token_dist}, reflecting long-tail problems' difficulties.

For multi-turn conversational datasets, we exclude the reasoning (\texttt{<think>}) content from all previous turns and retain only the reasoning content from the final turn, applying supervision solely on the last-turn output. This mirrors inference-time behavior, where earlier reasoning traces are never injected into subsequent turns and prevents uncontrolled context growth. In contrast, tool-calling datasets may require tool invocation at any turn, and thus include reasoning annotations for all turns. To ensure the model learns multi-step tool-use behavior correctly, loss is applied across every turn during training, even though inference-time tool executions similarly avoid propagating prior reasoning traces between turns.

\subsection{Ablations \& Key Findings}
\begin{figure}[t!]
\centering
\includegraphics[width=.49\textwidth]{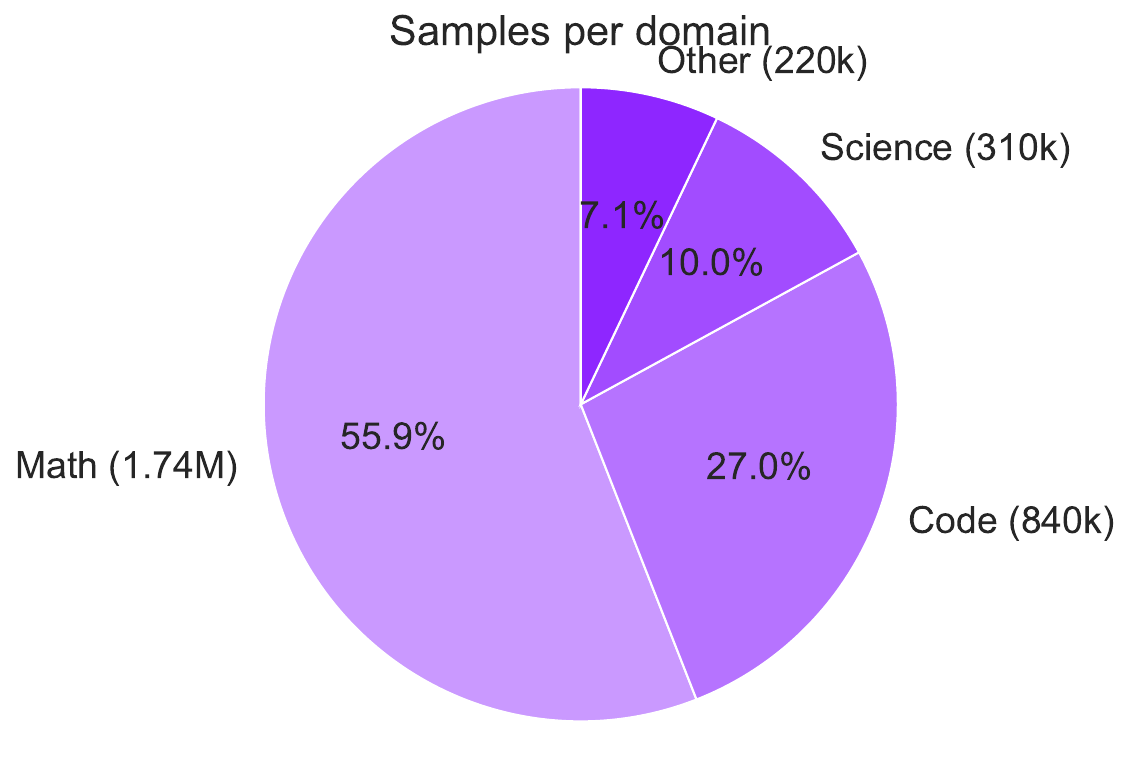}
\includegraphics[width=.49\textwidth]{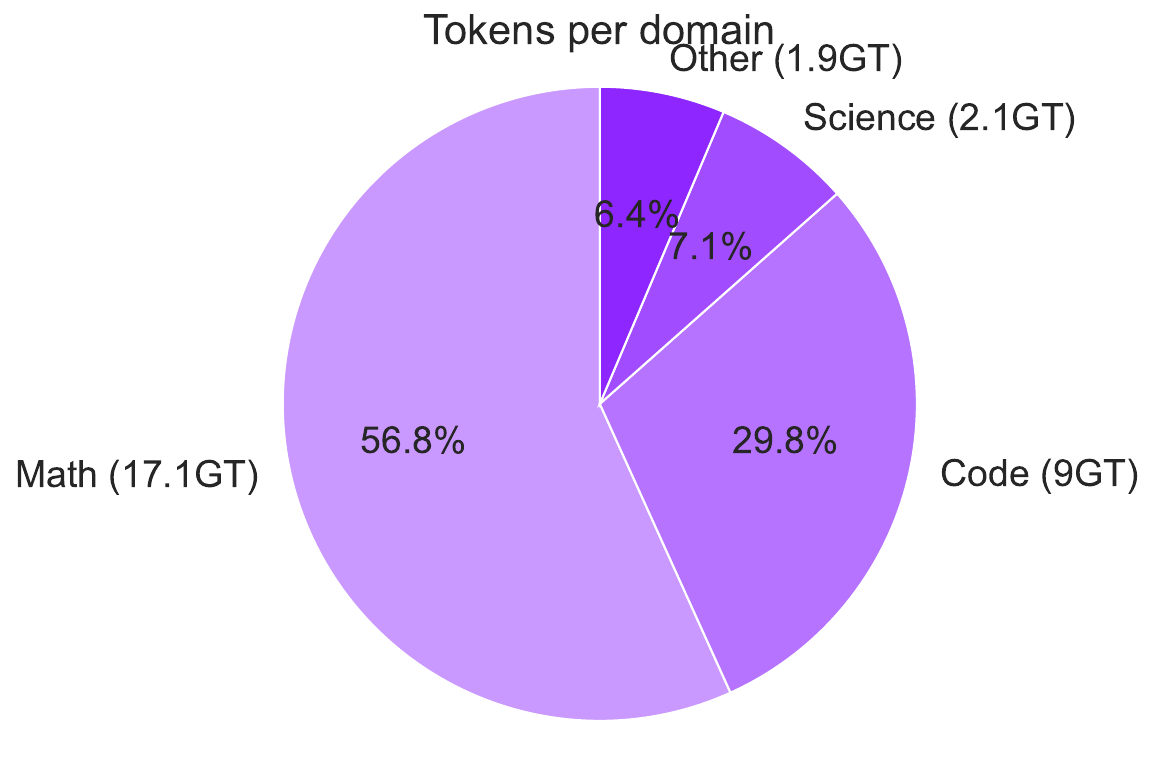}
\caption{Distribution of data categories in the SFT stage. The ``Other'' category includes IF, Chat, Safety, and Tool calling data.}
\label{fig:data_pie}
\end{figure}

To optimize our SFT stage, we conducted several ablations across the following axes: learning rate tuning, rollout multiplicity per input prompt, incorporation of incorrect rollouts, teacher model mixing, domain-specific data mixtures, and data weighting strategies. Below, we provide a detailed description of the conducted ablations and corresponding key findings:
\begin{itemize}
    \item \textbf{Learning rate tuning:} We selected the optimal learning rate through grid-searching the best validation performance on key reasoning benchmarks using a $10\%$ representative subset of our training corpus. We also leveraged $\mu P$ (\citep{falconH1}, Section 3.2.3) which notably ensures the transferability of the training hyper-parameters to the different model sizes of the Falcon-H1 series.
    \begin{tcolorbox}[
    colback=tiiPurple!5!white,
    colframe=tiiPurple!75!black,
    title=Use large LRs,
    fonttitle=\bfseries
    ]
    Larger learning rates consistently yielded superior model performance in our ablation runs (optimal selected LR was $1024\cdot10^{-6}$), outperforming smaller values typically recommended in standard SFT settings (e.g., for building instruct models). The selected LR has led to faster convergence as well as higher performance on downstream tasks.
    \end{tcolorbox}
    \item \textbf{Rollout multiplicity and correctness analysis:} We investigated the effect of varying the number of solution rollouts per problem instance, comparing $n\in \{ 4, 8, 12\}$. Ablation runs assessed whether increasing the number of diverse reasoning traces exposed the model to richer problem-solving strategies, particularly for complex problems. Additionally, we analyzed the impact of incorporating incorrect rollouts into the training set, evaluating in which conditions such traces would be beneficial.
    \begin{tcolorbox}[
    colback=tiiPurple!5!white,
    colframe=tiiPurple!75!black,
    title=Use high rollout counts,
    fonttitle=\bfseries
    ]
    Higher rollout counts ($n=12$) were most effective, especially for difficult problem queries. The performance gain scaled with problem difficulty: harder problems showed greater boosts from increased rollout diversity, suggesting that exposure to multiple valid reasoning traces is crucial for the model to acquire robust and generalizable problem-solving skills. On the other hand, including incorrect rollouts resulted only marginal gains on the hardest problems (i.e., with low baseline pass rates). For easier or intermediate problem difficulties, the inclusion of incorrect rollouts showed almost negligible gains and, in some cases, led to slightly noiser training signals, suggesting that their benefit is highly problem-dependent.
    \end{tcolorbox}
    \item \textbf{Teacher model mixing:} We studied the impact of mixing reasoning traces generated from different teacher models, using in-domain (e.g., math only) or cross-domains (e.g., math and code) seeds. Our initial hypothesis was that mixing outputs from different teachers would increase data diversity and potentially improve generalization.
    \begin{tcolorbox}[
    colback=tiiPurple!5!white,
    colframe=tiiPurple!75!black,
    title=Single-teacher outperformed multi-teachers,
    fonttitle=\bfseries
    ]
    Cross-domain teacher mixing was found to be counterproductive: training the model on data from multiple teacher reasoning traces exhibited higher output entropy and lower evaluation scores compared to the optimal single-teacher baseline. We hypothesize that conflicting reasoning styles across different teachers introduced distribution shifts and inconsistencies, yielding diminishing returns in model generalization. 
    \end{tcolorbox}
    \item \textbf{Domain-Specific Ablations and Difficulty-Aware Weighting:} To maximize performance within distinct reasoning domains (e.g., math, code, science), we conducted targeted ablations for each domain, by maximizing downstream performance on domain-specific data with corresponding key benchmarks. For designing the final mixture, we introduced a difficulty-aware weighting scheme: hard problems (as determined by pass rates or traces length) were treated as high-quality and up-weighted by factors between $1.25\times$ and $1.75\times$ to ensure sufficient model exposure, medium-difficulty samples retained standard weighting of $1$, and easy problems were progressively down-weighted $0.5\times$ or even removed in case of diminishing returns. This strategy focused training with harder problems yielding notable performance boosts, while minimizing overfitting to trivial samples. 
    \item \textbf{Full Training Mixture Ablations:} 
    Building on our domain-specific findings, we experimented with multiple data-mixture configurations. An initial mixture emphasized mathematics as the dominant component, with smaller contributions from code, science, and general reasoning. A refined mixture—used for the final model—increased the proportions of code and non-math reasoning data while slightly reducing the mathematical share. This produced a still math-dominant mixture but with broader cross-domain reasoning coverage and improved overall balance.
    \begin{tcolorbox}[
    colback=tiiPurple!5!white,
    colframe=tiiPurple!75!black,
    title=Math reasoning skills tend to transfer more to other domains,
    fonttitle=\bfseries
    ]
    The most effective training mixture was math-dominant with moderate inclusion of code and science data. This configuration outperformed more balanced or code-centric mixtures, suggesting that mathematical reasoning skills transfer more effectively to other domains than vice versa.
    \end{tcolorbox}
\end{itemize}
\noindent These ablation experiments provided empirical guidance for selecting the final set of SFT hyper-parameters and data-mixture composition. The final hyper-parameters and data-mixture configurations—selected based on the
full set of studies, ablations, and downstream evaluations—are summarized in Table \ref{tab:sft_hyperparams} and Figure \ref{fig:data_pie}.

\subsection{Distributed Training Setup}
The SFT stage for \textbf{Falcon-H1R-7B} training was performed using Fully Sharded Data Parallelism (FSDP) and context parallelism with $\text{CP}=2$, allowing efficient scaling to $256$ H100 GPUs of long-context finetuning. To enable Ulysses sequence parallelism~\citep{jacobs2023deepspeedulyssesoptimizationsenabling},
the Falcon-H1 hybrid Transformer--Mamba block was monkey-patched with explicit
\texttt{gather} and \texttt{scatter} operations. These modifications ensure that the block’s mixed attention and state-space computations execute correctly under sequence-partitioned parallelism. The default context length was set to $36$K tokens, with some few extended samples (up to $48$K tokens) right-trimmed to fit. The effective batch size was set to $512$, and we used a $4\%$ warmup phase of the total number of training steps. Training was performed over three epochs on a total of $3.1$M samples. Table~\ref{tab:sft_hyperparams} summarizes the main SFT hyperparameters.\\

\begin{table}[t!]
\centering
\small
\begin{tcolorbox}[
    colback=tiiPurple!3!white,
    colframe=tiiPurple!75!black,
    arc=6mm,
    boxrule=0pt,
    center,       
    left=0pt,
    right=0pt,
    top=4pt,
    bottom=4pt,
    width=.85\textwidth,
    sharp corners=downhill
]
\centering  
\color{tiiPurple!75!black}
\begin{tabular}{ll}
\textbf{Parameter} & \textbf{Value} \\
\hline\\[-1em]
Base Model & \href{https://huggingface.co/tiiuae/Falcon-H1-7B-Base}{Falcon-H1-7B-Base} \\
Parallelism Strategy & FSDP + Context Parallelism (CP=2) \\
Maximum Context Length & 36K tokens \\
Extended Context Samples & Up to 48K tokens (right-trimmed) \\
Number of GPUs (H100) & $256$ \\
Learning Rate & $1024\times10^{-6}$ with $\mu P$ scaling \citep{falconH1}\\
Learning Rate Scheduler & Linear \\
Batch Size & 512 \\
Warmup Ratio & 4\% of total steps \\
Epochs & 3 \\
Optimizer & AdamW (default $\beta_1$, $\beta_2$) \\
Weight Decay & 0.01 \\
Precision & bfloat16 \\
Gradient Clipping & 1.0 \\
\end{tabular}
\end{tcolorbox}
\caption{Hyper-parameters used for the Supervised Fine-Tuning (SFT) stage.}
\label{tab:sft_hyperparams}
\end{table}
To support scalable training, we enhanced the codebase with a streaming-dataset implementation and weighted data-mixture support. The streaming pipeline minimizes CPU memory usage and ensures that effective batch composition accurately reflects the intended mixture proportions. We further observed that introducing a periodic call to \texttt{torch.cuda.empty\_cache()} effectively mitigates memory fragmentation, preventing sporadic latency spikes and helping maintain consistent per-step training times during prolonged runs. To improve computational efficiency, we leverage \textit{Liger Kernels}, which provide fused and memory-optimized implementations for \textit{Rotary Position Embedding (RoPE)}, \textit{RMSNorm}, and \textit{Cross-Entropy} operations. These kernels enable significant memory savings and throughput improvements during training by reducing redundant memory reads and kernel launch overheads (details in Section \ref{sec:train-optim}).

\paragraph{Data-parallel Balance Tokens:} Supervised fine-tuning with long reasoning traces combined with short instruction-response pairs (as depicted in Figure \ref{fig:response_token_dist}) induces a bias when training with data parallelism (DP). In standard DP, each rank may process batches with widely varying numbers of valid tokens. When the loss is averaged \emph{locally} on each rank, every rank contributes an equally weighted loss to the global optimization step, irrespective of the actual number of valid tokens processed, which can yield large or noisy gradients from ranks processing short sequences. To overcome these imbalances, we employ a \textit{balanced data-parallel token normalization} strategy which has been proved to be effective in our ablations. Essentially, for each DP rank $r$, let $\ell_i^{(r)}$ denote the per-token loss on rank $r$, and $m_i^{(r)}\in \{0, 1\}$ be the corresponding loss mask indicating valid tokens. The \textit{data-parallel balanced loss} across each DP rank $r$ expresses as
\begin{align*}
    \mathcal{L}_{\text{balanced}}^{(r)} = \frac{ \sum_i \ell_i^{(r)} m_i^{(r)} }{ \varepsilon + \sum_{r=1}^R \sum_{i} m_i^{(r)} } \cdot R,
\end{align*}
where $\varepsilon$ stands for some small value for numerical stability. Notably, this balanced data-parallel token normalization  ensures that each token in the global batch contributes equally to the optimization step, regardless of sequence length. This is especially important for reasoning SFT, where training samples range from ultra-long chain-of-thought annotations and multi-step reasoning traces to short instruction-following examples. The procedure reduces gradient variance, stabilizes training, and prevents certain sequence lengths from being over- or under-weighted, resulting in more consistent and balanced training updates.\\

\begin{figure}[t!]
    \centering
    \includegraphics[width=.8\linewidth]{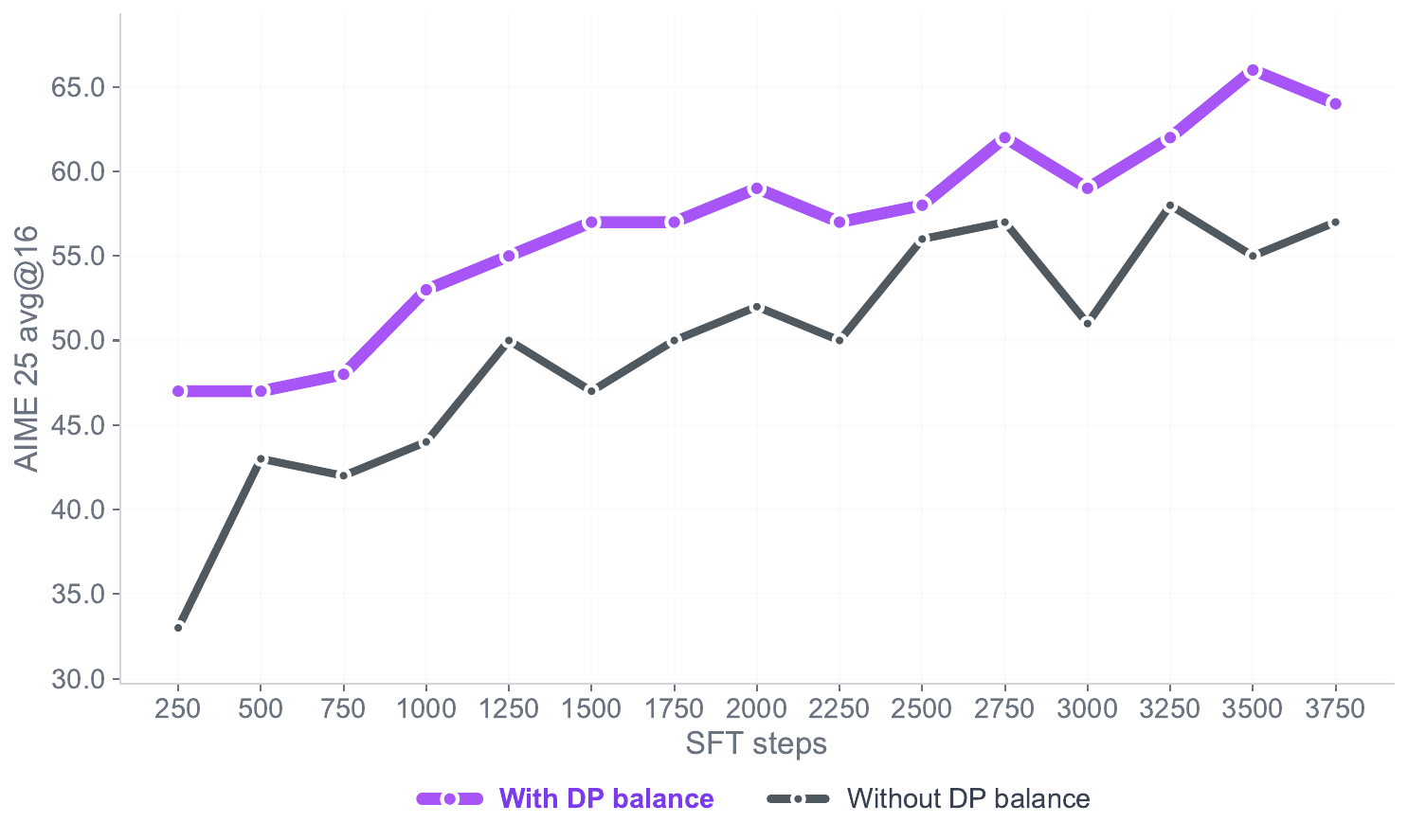}
    \caption{Effect of enabling \textit{Data-Parallel Balance} on downstream reasoning task.}
    \label{fig:dp_balance}
\end{figure}

\noindent Figure~\ref{fig:dp_balance} shows the performance gap on the AIME-25 benchmark during the first 3,750 SFT training steps of the Falcon-H1R-7B model, comparing models trained with and without Balanced Data-Parallel (DP) Token Normalization using our final data mixtures. This improvement is made possible by the \textsc{verl}~\citep{sheng2024hybridflow} update introduced in PR\footnote{\url{https://github.com/volcengine/verl/pull/3691}}, which enables correct aggregation of global token counts across data-parallel ranks. The impact is both substantial and consistent: models using Balanced DP Token Normalization achieve 4--10\% higher accuracy throughout training. These results demonstrate that equalizing per-token gradient contributions across heterogeneous batches, particularly when handling both long and short reasoning sequences, reduces gradient noise and stabilizes learning, leading directly to improved reasoning performance during SFT.

\section{Reinforcement Learning Stage}\label{sec:RL}
After completing the supervised fine-tuning (SFT) stage, we performed Reinforcement Learning with Verifiable Rewards (RLVR) to further enhance the reasoning capabilities of the model, i.e., improve pass@1\footnote{\citep{mroueh2025reinforcement} showed theoretically under a non-clipping GRPO objective that GRPO \textit{amplifies} the policy's probability of success, in line with the empirical findings of \citep{yue2025does}.}. Although the SFT model performed already well on reasoning benchmarks, the RLVR-stage further showed some extra boost in model's reasoning performance while simultaneously improving output quality, for example, by controlling response length. To achieve these goals, we developed a custom RL framework on top of the \textsc{verl}\footnote{\url{https://github.com/volcengine/verl}} library to perform GRPO-based RLVR. In the following subsections, we describe in detail our RL methodology, including data sources and curation techniques, training setup, and ablation studies conducted to produce the final model.

\subsection{Data Preparation}\label{sec:data-rl}
For the RL stage, we curated high-quality datasets focused on mathematics and coding. Each dataset was rigorously verified and filtered by difficulty to ensure effective learning of the model and control the distribution of problems' difficulties. To prevent memorization, the RL datasets are entirely disjoint from the SFT data, ensuring that improvements result solely from reward-based optimization rather than exposure to previously seen examples.
\paragraph{Data Curation:} To ensure consistent and reliable reward signals, ground-truth solutions were thoroughly verified across all domains before inclusion. For math, we excluded problems with answers that cannot be reliably verified by our rule-based checker. For coding, both curated and generated test cases were used to assess functional correctness and generalization across diverse programming tasks. We then filtered the training data based on problem difficulty estimated using different methods, based on data availability: (i) Publicly available difficulty metadata when provided. (ii) Pass-rate statistics when ground-truth solutions are available. (iii) Predicted difficulty from an LLM-based classifier otherwise.\\

\begin{wrapfigure}{r}{0.5\textwidth}
    \vspace{-10pt}
    \centering
    \includegraphics[width=.47\textwidth]{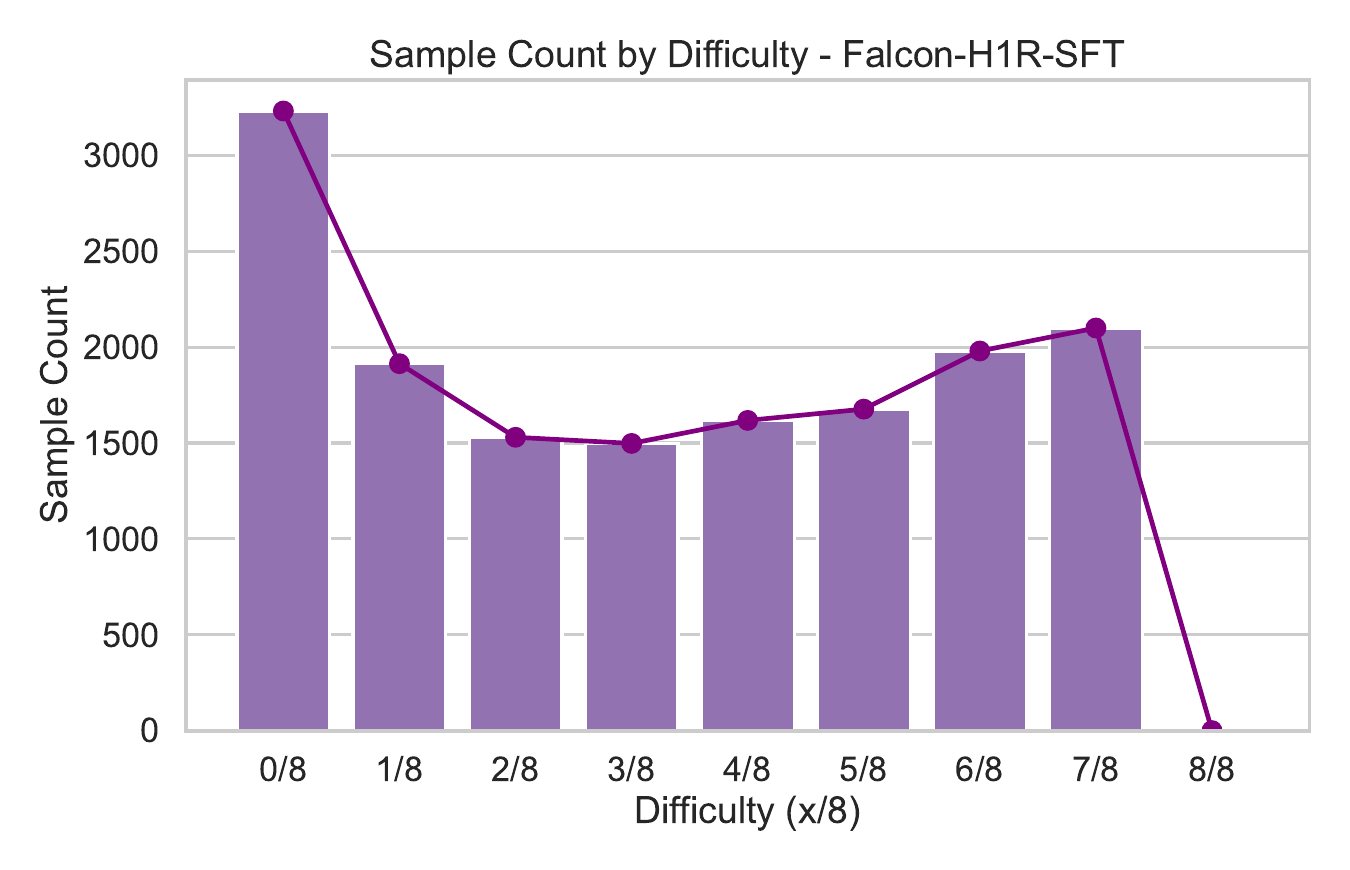}
    \caption{Math prompts difficulty distribution relative to the Falcon-H1R-SFT checkpoint.}
    \label{fig:7b_rl_difff}
    \vspace{-10pt}
\end{wrapfigure}

\paragraph{Data Processing:} As illustrated in Figure \ref{fig:rl-data-diagram}, the data processing stage for RL consisted of the following steps:

\begin{enumerate}
    \item \textit{Deduplication:} We first de-duped all the math and code data against the original SFT data to ensure no overlap between training stages.
    \item \textit{Difficulty Filtering:}
    To evaluate the difficulty of the curated dataset relative to the SFT model and filter it accordingly, we generated 8 solution rollouts for each problem using the sampling parameters ($\tau=0.6$, $p=0.95$, $\text{max-tokens}=32$K). The percentage of correct solutions served as a proxy for difficulty. The filtering process was performed as follows:
\begin{enumerate}[label=(\alph*)]
    \item \emph{Easy problems were removed:} Problems with $100\%$ success rate were discarded.
    \item \emph{Hard problems were undersampled:} For problems with $0\%$ success rate:
    \begin{enumerate}[label=(\roman*)]
        \item Problems where all solutions exceeded the token limit were majorly filtered out, and a small subset of truncated-solutions was retained through random sampling.
        \item For the remaining problems, the majority vote and its frequency were computed.
        \item Problems where the majority vote frequency $\geq 4$ were removed.
    \end{enumerate}
\end{enumerate}
\end{enumerate}

The final math dataset after the above filtering exhibited a mirrored J-shaped distribution similar to \citep{polaris} as depicted in Figure \ref{fig:7b_rl_difff}.

\subsection{Training Framework}

\paragraph{Algorithm:} Starting from the standard GRPO approach \citep{DSR1}, which corresponds to the following objective: 
\begin{align*}
J_{\text{GRPO}}(\theta)
= \frac{1}{G} \sum_{i=1}^{G} 
\frac{1}{|o_i|} \sum_{t=1}^{|o_i|}
\Bigg\{
\min\!\left[
\frac{\pi_{\theta}(o_{i,t} \mid q, o_{i,<t})}
     {\pi_{\theta_{\text{old}}}(o_{i,t} \mid q, o_{i,<t})}
A_{i,t},\;
\text{clip}\!\left(
\frac{\pi_{\theta}(o_{i,t} \mid q, o_{i,<t})}
     {\pi_{\theta_{\text{old}}}(o_{i,t} \mid q, o_{i,<t})},
1-\epsilon,\; 1+\epsilon
\right) A_{i,t}
\right]
\\
-\, \beta\, D_{\mathrm{KL}}(\pi_{\theta} \,\|\, \pi_{\text{ref}})
+\, \gamma\, \mathrm{Entropy}(\pi_{\theta})
\Bigg\}.
\end{align*}

\noindent
where the group-relative advantage, $A_{i,t}$ is defined as
\begin{align*}
A_{i,t}
=
\frac{
R_{\text{final}}(q, o_i)
-
\mathrm{mean}\bigl(
\{R_{\text{final}}(q, o_1), \ldots, R_{\text{final}}(q, o_G)\}
\bigr)
}{
\mathrm{std}\bigl(
\{R_{\text{final}}(q, o_1), \ldots, R_{\text{final}}(q, o_G)\}
\bigr)
}.
\end{align*}
and $D_{\text{KL}}$ is the KL-penalty term defined as:
\begin{align*}
D_{\text{KL}}(\pi_\theta \,\|\, \pi_{\text{ref}}) 
:= \frac{1}{G} \sum_{i=1}^{G} \frac{1}{|o_i|} \sum_{t=1}^{|o_i|} 
\frac{\pi_{\text{ref}}(o_{i,t} \mid q, o_{i,<t})}{\pi_\theta(o_{i,t} \mid q, o_{i,<t})}
- \log \frac{\pi_{\text{ref}}(o_{i,t} \mid q, o_{i,<t})}{\pi_\theta(o_{i,t} \mid q, o_{i,<t})}-1.
\end{align*}

We incorporated additional techniques, proven effective in the literature, to avoid training instability due to slow convergence, policy collapse, or training-inference mismatches. These adjustments targeted sampling strategies, loss functions, and computation efficiency for improved stability and convergence. The incorporated GRPO changes are detailed as follows:
\begin{itemize}
    \item \textit{Clip high, no KL-penaly and Entropy terms \citep{DAPO, drgrpo}:} Following \citep[Section 3.1]{DAPO}, we modify the upper importance sampling threshold. This adjustment encourages more exploratory token generation by preventing the policy from becoming overly conservative during training. Both the KL-penalty and entropy regularization terms
$\beta\, D_{\mathrm{KL}}(\pi_\theta \,\|\, \pi_{\theta_{\text{old}}}) \text{ and } \gamma\, \mathrm{Entropy}(\pi_\theta)$ are removed entirely by setting \(\beta = 0\) and \(\gamma = 0\). This allows deviations from the old policy thereby allowing more exploration. In our setting, keeping the exploration bonus $\gamma$ strictly positive was not necessary as the training with $\gamma = 0$ was relatively stable.
    \item \textit{Truncated Importance Sampling (TIS) \citep{yao}:} To reduce mismatch between sampling-time and training-time behavior, we incorporate the TIS correction, ensuring closer alignment between the model’s inference dynamics and its optimization updates.
    \item \textit{Cross Entropy Loss \citep{seed2025seed1}:} We add a cross-entropy loss term computed exclusively on positive (correct) samples. This auxiliary objective accelerates convergence by providing additional direct supervision signal while maintaining the benefits of the policy gradient approach.
    \item \textit{Online Sampling \citep{DAPO}:}  A challenge in GPRO training arises when entire batches contain zero-advantage samples—cases where all rollouts are uniformly correct or uniformly incorrect. Although these samples provide no policy-gradient signal, they still influence auxiliary losses such as the entropy bonus or supervised loss, which can destabilize training. To mitigate this, we evaluated three strategies:
    \begin{enumerate}
        \item \textit{Backfill}: Discard zero-advantage prompts and iteratively draw additional samples from later batches until the target batch size is met.
        \item \textit{Reject}: Perform strict rejection sampling, allowing the batch size to vary depending on the number of valid samples.
        \item \textit{Duplicate}: Randomly duplicate positive-advantage samples to restore the batch size after filtering.
    \end{enumerate}
    In our ablations, we found that the \textit{backfill} approach provided the most stable performance, which we then adopted for our final setup. To improve the efficiency of the backfill strategy, we introduced a generation caching system with a bounded buffer in the same spirit as \citep{kimi}. This mechanism addresses a key inefficiency: when additional samples are needed to complete a batch, standard backfill often generates excess rollouts that are discarded, wasting computation. Our cache stores these surplus prompts for future use, with a fixed size to prevent staleness and optional periodic refreshes to ensure samples remain relatively on-policy. This approach reduces generation calls by approximately 30\% while maintaining the freshness of training data through careful cache size limits and periodic refreshing.
\end{itemize}

\paragraph{Reward Functions Design:} We employ different reward schema for each domain: math reasoning, code generation and science reasoning. The reward orchestration layer dynamically selects the appropriate reward schema based on the task domain. Reward computation is executed in a distributed manner using Ray with AsyncIO coordination, while ThreadPoolExecutor accelerates I/O CPU-bound workloads.
\begin{itemize}
    \item \textit{Math Reward:} For mathematical tasks we use a binary reward system based on the correctness of the final extractable answer within the response. We employ a two-tier verification system wherein the first tier extracts and validates answers using \textsc{LaTeX} string matching and symbolic equivalence checks for fast heuristics. If tier one marks a response as incorrect, a semantic verification fallback using \textsc{math-verify}\footnote{\url{https://github.com/huggingface/Math-Verify}} is triggered to handle more complex ground truths and provide additional validation.

    \item \textit{Code Reward:} The execution backend for the code reward system is based on  \textsc{Sandbox-Fusion}\footnote{\url{https://github.com/bytedance/SandboxFusion}}. To maximize concurrency and reduce reward evaluation latency, we deploy multiple instances of the Sandbox-Fusion execution engine using Docker containers. The main execution flow of the reward design is as follows:
    \begin{enumerate}
        \item \textit{Code extraction and language detection:} Locate fenced code blocks using regex and detect programming language (Python, Java, C++, C, JS, Bash, Go, etc.).
        \item \textit{Sample test cases:} To ensure efficient and scalable evaluation across the different sandboxes, we randomly select up to a configurable number of test cases from each data sample. 
        \item \textit{Execute in Sandbox-Fusion:} The payload is submitted via an \textsc{HTTP} request to the sandbox API in a round-robin dispatch across distributed sandbox instances. Just as for math, we use Boolean pass or fail reward with reward $1.0$ assigned only in the event all tests pass. 
    \end{enumerate}
    
    \item \textit{Science Reward:} We also implemented a reward system using an LLM served through parallel \texttt{vLLM} instances to evaluate output correctness, especially in cases where traditional math-based reward functions struggle with complex ground truths, such as in science problems. The LLM judge takes the prompt, extracted answer, and ground-truth solution as input and produces a single scalar value that indicates whether the response is correct.
    
    \item \textit{Format Reward:} Additionally, we penalize incoherent outputs and the use of mixed languages by applying a format-specific penalty for responses that do not adhere to the \texttt{<think> CoT </think> Answer} structure.

\end{itemize}

\subsection{Training Ablations}
We conducted two forms of ablation studies. First, we explored various GRPO hyperparameters—including rollout sampling strategy, generation length, reward model configuration, and other optimization settings—to identify the optimal configuration for our training framework. Second, we investigated the best training strategy for mixing multiple data domains by experimenting with a range of RL setups, e.g., feeding data domains sequentially in multiple RL stages versus a combined RL stage with mixing data domains. Below, we summarize the selected hyperparameters (see Table \ref{tab:rl_hyperparams}) and present the results for each training setup.

\paragraph{Selecting Optimal Hyper-parameters:} The RL training hyperparameters were selected to ensure maximum diversity among rollouts and to manage the model's response length. We summarize below the key choices.
\begin{itemize}
    \item \textit{Group Size (G):} We experimented with different group sizes, $G \in \{ 4, 8, 16, 32\}$. Increasing the number of rollouts from $G = 4$ to $G = 16$ improved response diversity and led to better performance on our downstream evaluations. However, further increasing to $G = 32$ did not yield significant gains and substantially increased generation time.
    \item \textit{Max Response Length ($L_{\text{max}}$):} Given the large response lengths generated by the SFT model, $L_{\text{max}}$ proved to be an important parameter for maximizing RL performance. We experimented with both relatively shorter maximum lengths, $L_{\text{max}}=24k$ tokens, and much longer lengths, $L_{\text{max}}=48k$ tokens. Ultimately, we found that, similar to the conservative approach adopted by \citep{polaris}, gradually allowing the model to \textit{think longer} from the beginning by setting $L_{\text{max}}= 48k$ was best suited for our SFT model and led to the highest performance.
    \item \textit{Sampling Temperature ($\tau$):} The training temperature was selected to balance two competing objectives: maintaining sufficient diversity in generated rollouts to encourage exploration, while preserving strong model performance. Following the methodology in \citep{polaris}, we conducted a systematic evaluation of the SFT model's accuracy across different sampling temperatures on the AIME24 benchmark. For training, we selected a temperature within the Controlled Exploration Zone as described in \citep{polaris} to be a transitional region where accuracy begins to decline moderately while diversity increases substantially.
    \item \textit{Entropy Control ($\gamma$):} After tuning $G$ and $\tau$ to achieve sufficient rollout diversity, we observed that the policy entropy remained stable without explicit entropy regularization. Therefore, we set $\gamma = 0$.
    \item \textit{KL coefficient ($\beta$):} In line with the findings in \citep{DAPO}, we removed the KL divergence term to allow the model to explore more freely beyond the limits of the original SFT model.
\end{itemize}

\begin{table}[t!]
\centering
\small
\begin{tcolorbox}[
    colback=tiiPurple!3!white,
    colframe=tiiPurple!75!black,
    arc=6mm,
    boxrule=0pt,
    center,       
    left=0pt,
    right=0pt,
    top=4pt,
    bottom=4pt,
    width=.9\textwidth,
    sharp corners=downhill
]
\centering  
\color{tiiPurple!75!black}
\begin{tabular}{ll}
\textbf{Parameter} & \textbf{Value} \\
\hline\\[-1em]
Clip ratio, ($\epsilon^{-} , \epsilon^{+}$)  & $(0.2,0.28)$ \\
Group size ($G$) & $16$  \\
Sampling temperature ($\tau$) &  $0.85$ \\
Number of rollouts ($G$)  & 16 \\
Maximum rresponse length ($L_{\text{max}}$) & $48K$ \\
Batch size & 128 \\
PPO batch size & 128 \\
Learning rate & $2\times 10^{-6}$ with $10$ steps of linear warmup phase \\
Entropy control ($\gamma$) & $0$\\
KL coefficient ($\beta$)   & $0$\\
TIS max cap & 2 \\
\end{tabular}
\end{tcolorbox}
\caption{Hyperparameters used for the RL stage.}
\label{tab:rl_hyperparams}
\end{table}

\paragraph{Selecting task domains:} To determine the optimal training setup and data for our RL experiments, we conducted a series of ablation studies to assess how different task-domain sampling strategies influence model performance. We evaluated mathematical reasoning using the AIME25 benchmark, code generation using LCB v6, and scientific reasoning using GPQA-Diamond. All experiments were run until performance gains plateaued, and we report the best observed results in Table \ref{tab:ablations}. Our analysis focused on the following experimental strategies:
\begin{itemize}

\item \textbf{Math-Only:} This ablation was conducted exclusively on math-reasoning problems, with rewards computed using the math reward function. It served as our primary setting for understanding model behavior under RL and for tuning core hyperparameters.
 \begin{tcolorbox}[
    colback=tiiPurple!5!white,
    colframe=tiiPurple!75!black,
    title=Math Only training leads to strong reasoning and good generalization,
    fonttitle=\bfseries
    ]
    Consistent with our SFT findings, incorporating math-focused data during RL was essential for achieving strong accuracy gains across reasoning-heavy benchmarks. Math remained the dominant contributor to downstream performance for our training setup.
    \end{tcolorbox}

\item \textbf{Code-Only:} To isolate performance improvements relevant to code generation, we trained solely on code-reasoning tasks using the code-sandbox reward model. This experiment aimed to assess the upper bound of code-centric performance achievable under our RL setup.

 \begin{tcolorbox}[
    colback=tiiPurple!5!white,
    colframe=tiiPurple!75!black,
    title=Code only training is strong on code but has weaker generalization,
    fonttitle=\bfseries
    ]
    Pure code-only training achieved the highest scores on code-centric benchmarks; however, this came at the expense of math and science performance, with generalization under code-only training being substantially weaker than that of math-only training.
    \end{tcolorbox}

\item \textbf{Science-Only:} To evaluate whether science-oriented training improved scientific reasoning for our setup, we trained exclusively on science tasks scored by an LLM-based evaluator. RL using an LLM-as-a-judge reward model for science tasks did not yield meaningful improvements on GPQA-Diamond, suggesting either limited coverage of science capabilities or insufficient signal quality from the LLM evaluator.

\item \textbf{Code Followed by Math:} The two-stage curriculum took the checkpoint from Math-Only training and added a Code-Only stage on top. The goal was to assess whether gains from code training compound when followed by math-focused refinement. Prior to training, we re-evaluated the difficulty distribution of our code dataset relative to the Stage 2 checkpoint. This analysis revealed a significant shift in distribution compared to the SFT baseline: many problems that were previously unsolvable by the SFT model became solvable after the two stages of math-focused RL training.

 \begin{tcolorbox}[
    colback=tiiPurple!5!white,
    colframe=tiiPurple!75!black,
    title=Sequential training provides modest gains for more compute,
    fonttitle=\bfseries
    ]
   Across the main math and code reasoning benchmarks, the sequential curriculum—math training followed by code training—achieved the best average performance across the benchmarks in Table \ref{tab:ablations}, although the gains were more modest.
\end{tcolorbox}

\item \textbf{Math and Code Combined:} In this mixed-domain setting, each batch contained a blend of math and code problems. The ablation was done to test whether joint training offered complementary benefits compared to sequential or single-domain approaches. In our training setup, this strategy did not outperform the alternatives, suggesting that domain interleaving offers only a limited advantage over domain-focused training.

\end{itemize}

\begin{table}[t!]
\begin{tcolorbox}[
    colback=tiiPurple!3!white,
    colframe=tiiPurple!75!black,
    arc=6mm,
    boxrule=0pt,
    center,       
    left=0pt,
    right=0pt,
    top=4pt,
    bottom=4pt,
    width=0.8\textwidth,
    sharp corners=downhill
]
\centering 
\small
\color{tiiPurple!75!black}
\begin{tabular}{lccc}
\textbf{Ablation} & \textbf{AIME25} & \textbf{LCB v6} & \textbf{GPQA-Diamond} \\
\hline\\[-1em]
SFT Baseline                 & 79.6 & 64.6 & 60.4  \\
Math Only                 & 83.1 & 68.6 & 61.3  \\
Code Only                 & 79.2 & 70.8 & 59.7 \\
Science Only              & 77.7 & 63.7  & 60.9 \\
Math $\rightarrow$ Code   & 82.1 & 69.6 & 62.2 \\
Math + Code (Mixed)       & 79.7 & 67.1 & 61.4 \\
\end{tabular}
\end{tcolorbox}
\caption{RL ablation results across different task domains.}
\label{tab:ablations}
\end{table}

\begin{figure}[t]
    \centering
    \begin{subfigure}[b]{0.45\textwidth}
        \centering
        \includegraphics[width=\linewidth, height=7.5cm, keepaspectratio]{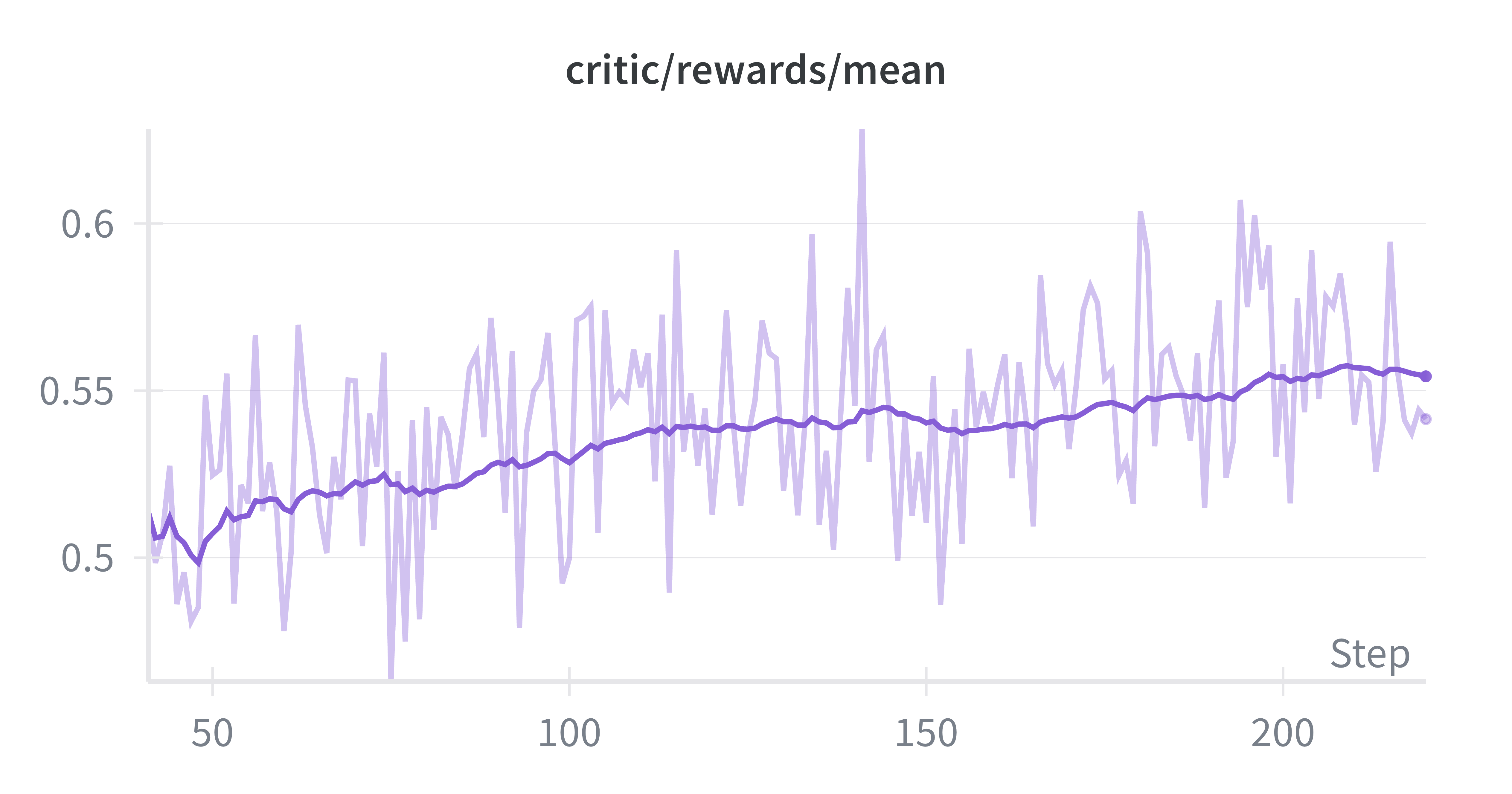}
        \caption{}
    \end{subfigure}
    \hfill
    \begin{subfigure}[b]{0.45\textwidth}
        \centering
        \includegraphics[width=\linewidth, height=7.5cm, keepaspectratio]{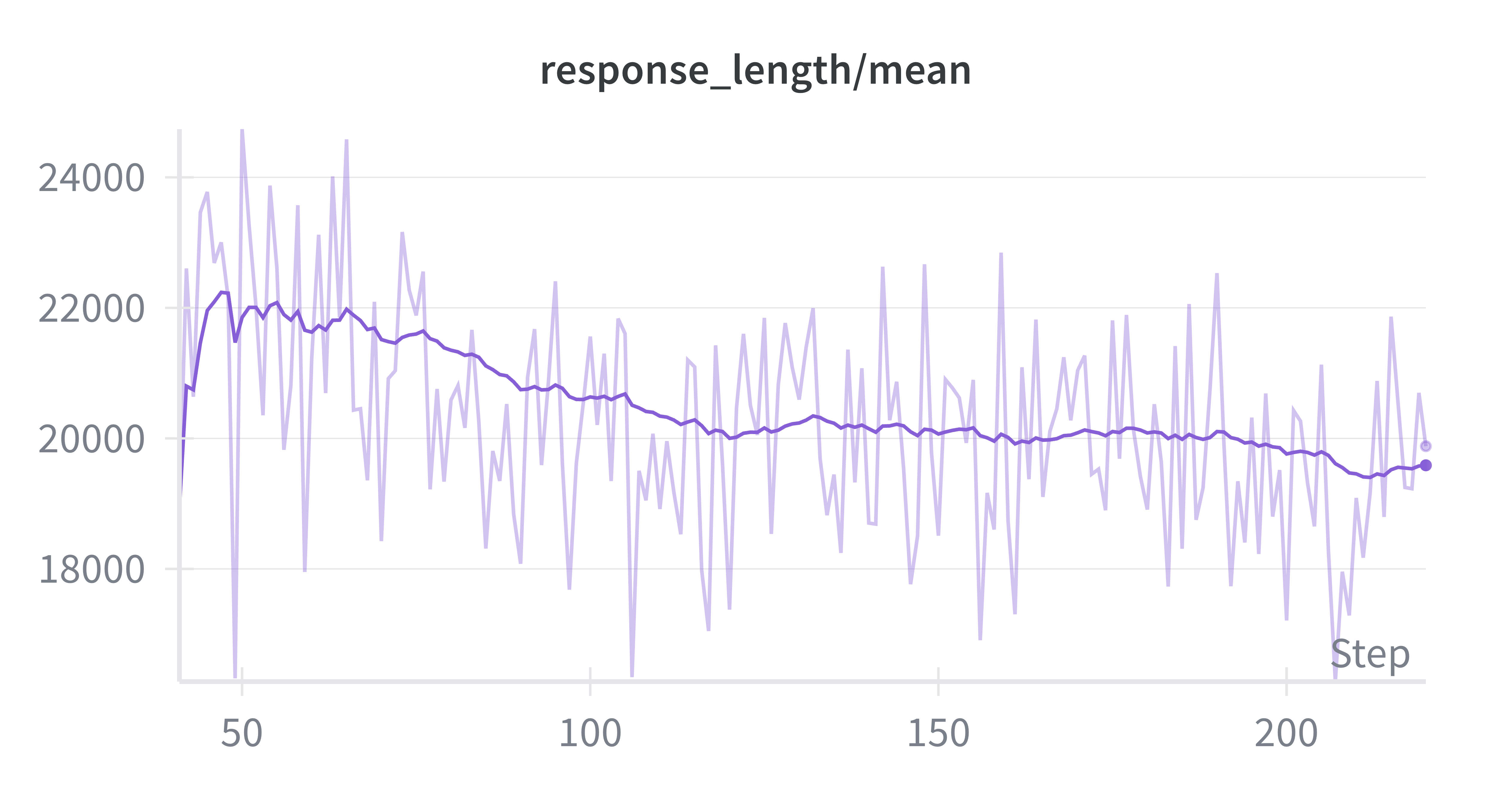}
        \caption{}
    \end{subfigure}
    
    \vspace{1em}
    
    \begin{subfigure}[b]{0.45\textwidth}
        \centering
        \includegraphics[width=\linewidth, height=7.5cm, keepaspectratio]{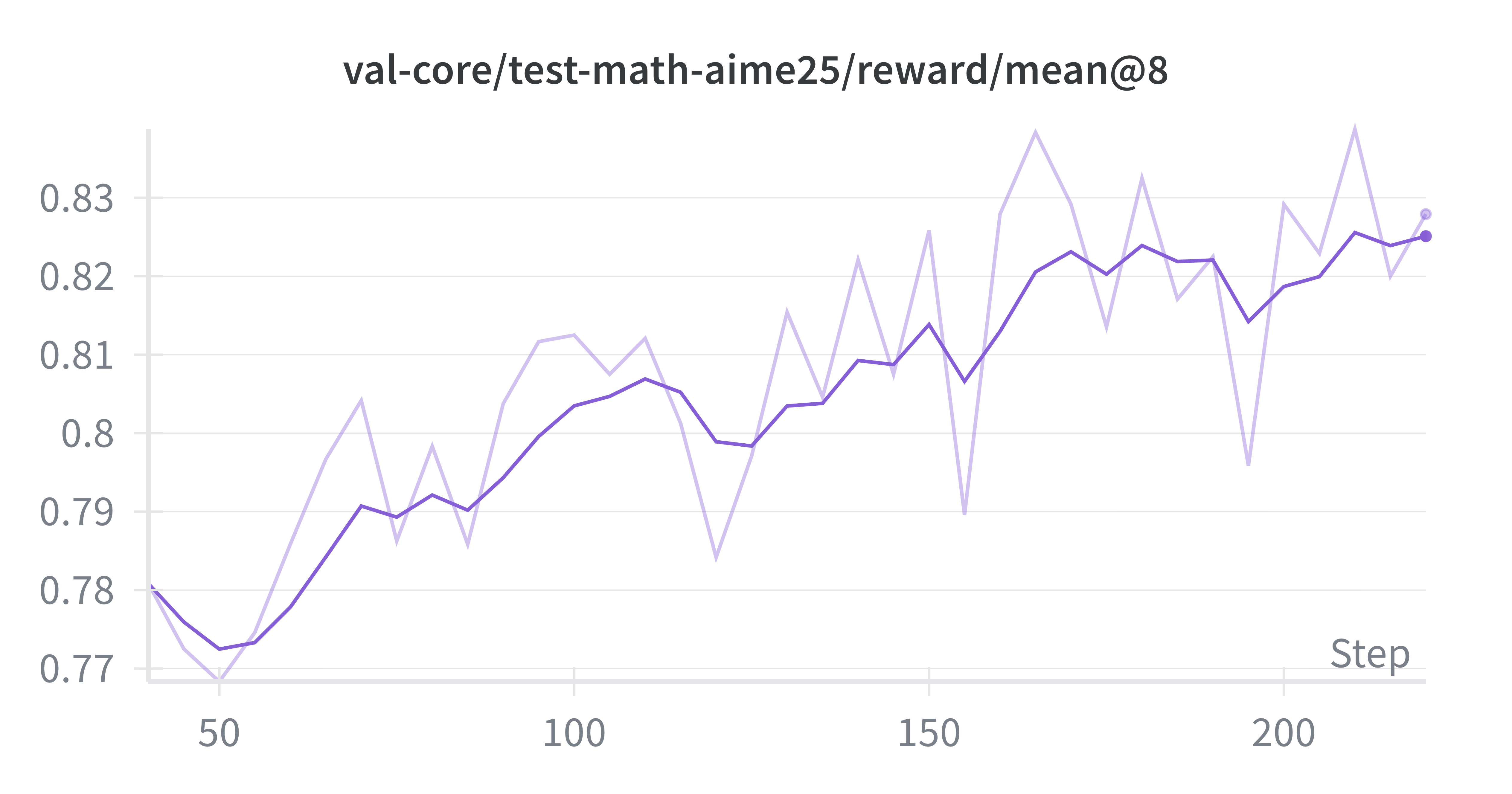}
        \caption{}
    \end{subfigure}
    \hfill
    \begin{subfigure}[b]{0.45\textwidth}
        \centering
        \includegraphics[width=\linewidth, height=7.5cm, keepaspectratio]{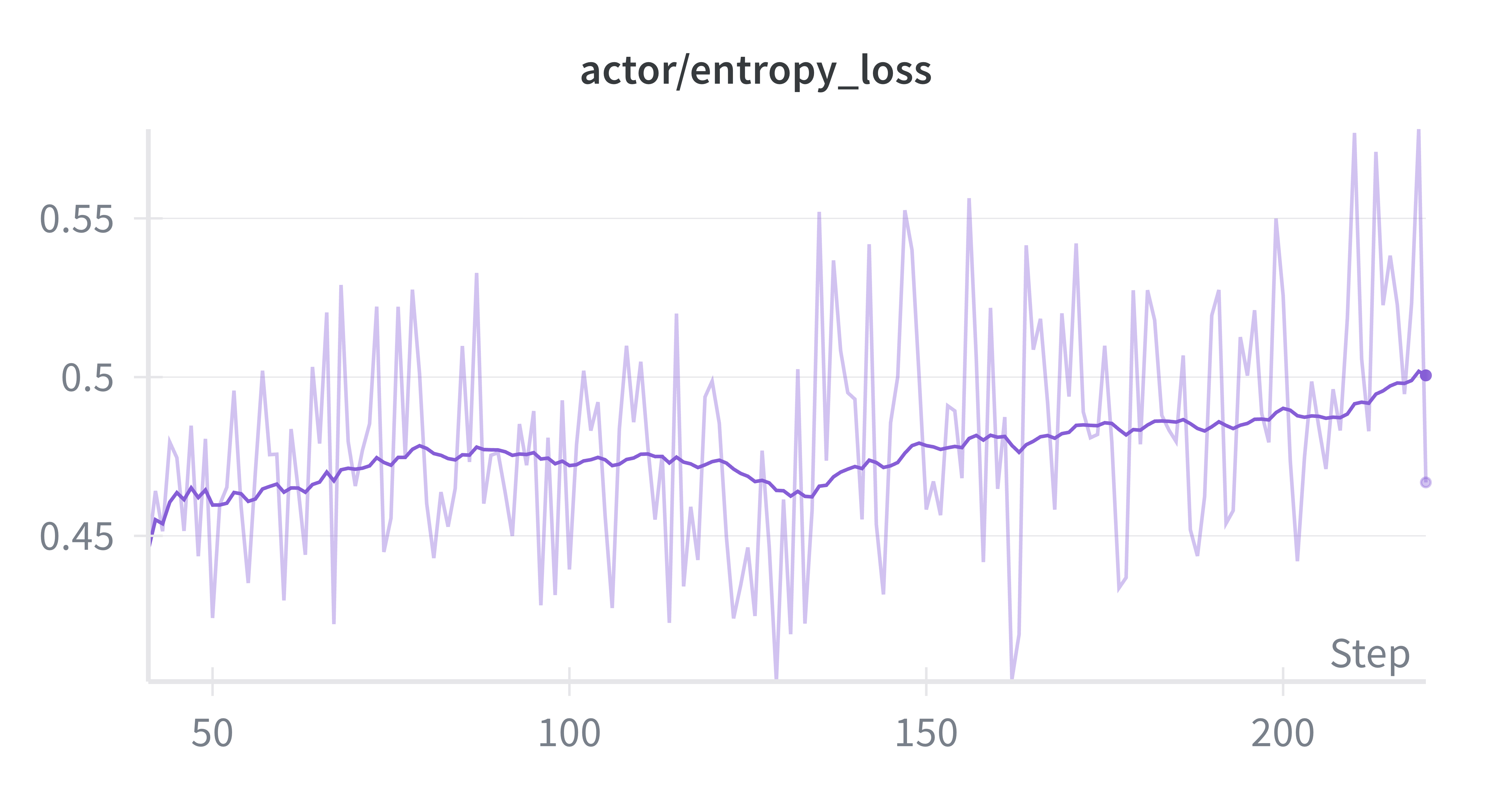}
        \caption{}
    \end{subfigure}
    \caption{Final RL training curves.}
    \label{fig:final-training-setup}
\end{figure}

\subsection{Final Training Setup}
Our final \textbf{Falcon-H1R-7B} RL training was performed using the hyperparameters in Table \ref{tab:rl_hyperparams}. The training was conducted on $256$ H100 GPUs using our custom GRPO framework built on \textsc{verl}. The curriculum consisted of a single \textbf{Math-Only} stage aimed at progressively enhancing reasoning capabilities while carefully managing sequence length, domain specialization, and exploration efficiency. We chose this single-stage design because, according to our ablation studies (see Table \ref{tab:ablations}), although adding a separate code stage yielded marginal improvements on the main benchmarks, it led to a decrease in average performance across the broader benchmarks we report in Section \ref{sec:eval}.\\

Prior to training, as illustrated in Figure \ref{fig:7b_rl_difff}, we filtered the mathematics dataset by difficulty using the pass rate from the SFT checkpoint. Training was then initialized with this curated set of mathematics-only data. We set the maximum response length to $48K$ tokens and used a sampling temperature of $\tau = 0.85$. This higher temperature encouraged more diverse reasoning during rollouts, allowing the model to explore a broader range of solutions. Additionally, the following system prompt was applied during training.

\begin{tcolorbox}[
    colback=tiiPurple!5!white,
    colframe=tiiPurple!75!black,
    title=System Prompt,
    fonttitle=\bfseries
]
You are Falcon, a helpful AI assistant created by Technology Innovation Institute (TII). To answer the user's question, you first think about the reasoning process and then provide the user with the answer. The reasoning process is enclosed within \texttt{<think> </think>} tags, i.e., \texttt{<think>} reasoning process here \texttt{</think>} answer here.
\end{tcolorbox}

\paragraph{Training Dynamics and Monitoring:} Throughout the training run, we monitored several key metrics to ensure stable optimization. The mean reward showed consistent improvement (Figure~\ref{fig:final-training-setup}a), while average response length stabilized at 20k as the model learned to refined its reasoning (Figure~\ref{fig:final-training-setup}b) across Stage 1. The entropy loss remained stable across training (Figure~\ref{fig:final-training-setup}d), suggesting healthy exploration without collapse.  The actor's policy-gradient clipping fraction remained below 1\%, indicating that updates were rarely clipped and optimization dynamics remained smooth.

\section{Standard Reasoning Tasks Evaluation}\label{sec:eval}
This section introduces the reasoning benchmarks and the evaluation methodology and results to assess the performance of \textbf{Falcon-H1R-7B}. Additional safety evaluations of our model are presented in Appendix \ref{sec:safety}.

\subsection{Benchmarks \& Methodology}
We evaluated our model using a diverse set of challenging reasoning benchmarks, organized into three categories: Math (AIME24, AIME25, HMMT25, AMO-Bench, MATH500), Code (LiveCodeBench v6, SciCode, $\tau^2$-Telecom, Terminal Bench Hard), and General (GPQA-Diamond, MMLU-Pro, Humanity Last Exam, IFBench). For convenience, brief descriptions of each benchmark are presented in Table~\ref{tab:benchmarks}. In our evaluations, we report pass@1 with the following settings:
\begin{tcolorbox}[
    colback=tiiPurple!3!white,
    colframe=tiiPurple!75!black,
    arc=6mm,
    boxrule=0pt,
    center,       
    left=0pt,
    right=0pt,
    top=4pt,
    bottom=4pt,
    width=.85\textwidth,
    sharp corners=downhill
]
\centering 
\small
\color{tiiPurple!75!black}
\begin{tabular}{ccc}
Responses per query &
Max response Length &
Benchmarks \\
\hline\\[-1em]
16 & 65536 & AIME24/25, AMO-Bench, HMMT25  \\
5 & 32768 & GPQA-Diamond \\
3 & 65536 & LiveCodeBench v6 \\
3 & 32768 & $\tau^2$-Telecom, TB Hard, MMLU-Pro \\
1 & 65536 & SciCode, HLE \\
1 & 32768 & MATH500, IFBench
\end{tabular}
\end{tcolorbox}
\noindent For our model's evaluation, we used the system prompt described in the previous section and set the sampling parameters to a temperature of $0.6$ and a Top-p value of $0.95$. In particular,
\begin{itemize}
    \item For MMLU-Pro, Humanity’s Last Exam (HLE), and GPQA-Diamond, we adopt the prompts and answer extraction regular expressions provided by Artificial Analysis (AA)\footnote{\url{https://artificialanalysis.ai/}}. Consistent with AA’s methodology, we use OpenAI's GPT-4o as the equality checker for HLE. 
    \item Terminal-Bench (TB) Hard evaluates 47 ‘hard’ tasks from the terminal-bench-core dataset (commit 74221fb) using the Terminus 2 agent. Each task is capped at 100 episodes and a 24-hour timeout, with model configurations set to the framework’s default values. 
    \item For $\tau^2$-Bench Telecom, each task is limited to 100 steps, and the Qwen3 235B A22B 2507 (Non-reasoning) model acts as the user agent simulator. 
    \item SciCode is evaluated using InspectAI\footnote{\url{https://inspect.aisi.org.uk}} as recommended by the SciCode repository\footnote{\url{https://github.com/scicode-bench/SciCode}}, and correctness is measured at both sub-problem and full-problem levels. 
    \item For HLE, we evaluate 2,158 text-only problems. For TB Hard, $\tau^2$-Bench Telecom, and SciCode, we repeat each evaluation three times and report the average result.
\end{itemize}

To compare performance, we evaluated a range of state-of-the-art (SOTA) models across 7B-32B parameter ranges, using each baseline's recommended evaluation configurations. For the 7B category, we included leading models such as Qwen3-8B \citep{qwen2025qwen3} and DeepSeek-R1-0528-Qwen3-8B \citep{DSR1}. In the mid-size range, we evaluated Phi-4-Reasoning-Plus-14B \citep{Phi4-Reasoning-Plus}, Apriel-1.5-15b-Thinker \citep{radhakrishna2025apriel1515bthinker}, GPT-OSS-20B \citep{openai2025gptoss120bgptoss20bmodel}, and Qwen3-32B \citep{qwen2025qwen3}. We also included Nemotron-H-47B-Reasoning \citep{nemotronh}, which represents the current leading hybrid reasoning model. 

\paragraph{Contamination analysis:} To ensure the integrity of our evaluations, we thoroughly analyzed our training datasets for contamination across all considered benchmarks. Our results revealed extremely low contamination levels. Specifically, exact string matching indicated $0\%$ contamination for all benchmarks except MMLU-Pro, which showed near-zero levels: $0.0005\%$ for our SFT mixture and $0.035\%$ for the RL data. 

\subsection{Evaluation Results}
Evaluation results for Math, Code, and General benchmarks are shown in Tables \ref{tab:eval-math}, \ref{tab:eval-code}, and \ref{tab:eval-general}, respectively. \textbf{Falcon-H1R} demonstrates competitive performance across all tasks, matching or exceeding state-of-the-art reasoning models, despite having a significantly smaller parameter count.

\paragraph{Mathematical Reasoning:} Falcon-H1R demonstrates exceptional mathematical reasoning capabilities, achieving the highest scores on AIME24 (88.1\%), HMMT25 (64.9\%), AMO-Bench (36.3\%), and MATH500 (97.4\%), and second place on AIME25 (83.1\%). Notably, our model surpasses much larger models, including Qwen3-32B, Nemotron-H-47B-Reasoning, and GPT-OSS-20B. On the AMO-Bench, which features advanced mathematical olympiad problems, Falcon-H1R scores 36.3\%, exceeding the next best model (GPT-OSS-20B) by over 10 percentage points. These results indicate that our training methodology enables strong generalization on complex, multi-step mathematical reasoning tasks.

\begin{table}[H]
\begin{tcolorbox}[
    colback=tiiPurple!3!white,
    colframe=tiiPurple!75!black,
    arc=6mm,
    boxrule=0pt,
    center,       
    left=0pt,
    right=0pt,
    top=4pt,
    bottom=4pt,
    width=\textwidth,
    sharp corners=downhill
]
\centering 
\small
\color{tiiPurple!75!black}
\begin{tabular}{lccccc}
\textbf{Models} &
\textbf{AIME24} &
\textbf{AIME25} &
\textbf{HMMT25} &
\textbf{AMO-Bench} &
\textbf{MATH500} \\
\hline\\[-1em]
Qwen3-8B & 77.9 & 65.8 & 41.0 & 14.1 & 97.4 \\
DeepSeek-R1-0528-Qwen3-8B & 83.3 & 75.8 & 54.3 & 23.3 & 96.8 \\
Phi-4-Reasoning-Plus-14B & 77.2 & 71.2 & 47.7 & 15.0 & 95.4 \\
Apriel-1.5-15b-Thinker & 86.2 & 80.0 & 61.0 & 22.2 & 97.2 \\
GPT-OSS-20B & 83.3 & 84.4 & 64.8 & 26.0 & 94.8 \\
Qwen3-32B & 79.4 & 71.0 & 49.8 & 21.3 & 96.8 \\
Nemotron-H-47B-Reasoning & 64.6 & 51.4 & 34.2 & 7.0 & 91.4 \\ 
\textbf{Falcon-H1R-7B} & \textbf{88.1} & \underline{83.1} & \textbf{64.9} & \textbf{36.3} & \textbf{97.4} \\
\end{tabular}
\end{tcolorbox}
\caption{Mathematical reasoning evaluations. \textbf{Bold} and \underline{underline} highlight top-1 and top-2 standings of Falcon-H1R-7B w.r.t.\@ the considered baselines.}
\label{tab:eval-math}
\end{table}

\paragraph{Code Generation:} On code benchmarks, Falcon-H1R attains the second-highest score on LiveCodeBench v6 (68.6\%), surpassed only by GPT-OSS-20B. Its performance on SciCode and domain-specific benchmarks (TB Hard, $\tau^2$-Telecom) remains competitive with models of similar scale.

\begin{table}[H]
\begin{tcolorbox}[
    colback=tiiPurple!3!white,
    colframe=tiiPurple!75!black,
    arc=6mm,
    boxrule=0pt,
    center,       
    left=0pt,
    right=0pt,
    top=4pt,
    bottom=4pt,
    width=\textwidth,
    sharp corners=downhill
]
\centering 
\small
\color{tiiPurple!75!black}
\begin{tabular}{lcccc}
\textbf{Models} &
\textbf{LCB v6} &
\textbf{SciCode sub / main} &
{\textbf{$\tau^2$-Telecom}} & {\textbf{TB Hard}} \\
\hline\\[-1em]
Qwen3-8B & 53.0 & 28.3 / 6.7 & 27.8 & 2.1 \\
DeepSeek-R1-0528-Qwen3-8B & 57.2 & 22.2 / 2.6 & 0.0 & 1.4 \\
Phi-4-Reasoning-Plus-14B & 53.1 & 29.8 / 7.2 & 0.0 & 2.1 \\
Apriel-1.5-15b-Thinker & 53.0 & {31.9} / {8.2} & 68.4$^{\ast}$ & 9.9$^{\ast}$ \\ 
GPT-OSS-20B & 72.0 & {34.9} / {6.2} & {60.2}$^{\ast}$ & {9.9}$^{\ast}$ \\
Qwen3-32B & 61.0 & 36.4 / 9.2 & 29.8 & 2.8 \\
Nemotron-H-47B-Reasoning & 47.4 & {26.1} / {4.6} & {11.4} & {1.4} \\ 
\textbf{Falcon-H1R-7B} & \underline{68.6} & 28.3 / 3.9 & 25.4 & \underline{4.9} \\
\end{tabular}
\end{tcolorbox}
\caption{Code generation with reasoning evaluations. \textbf{Bold} and \underline{underline} highlight top-1 and top-2 standings of Falcon-H1R-7B w.r.t.\@ the considered baselines. Results marked with $^*$ are taken from Artificial Analysis.}
\label{tab:eval-code}
\end{table}

\paragraph{General Reasoning:} Falcon-H1R also performs strongly on general benchmarks, achieving second-best results on HLE (11.1) and IFBench (53.4). The IFBench score highlights the model’s robust instruction-following capabilities, an important attribute for practical deployment. While results on GPQA-Diamond and MMLU-Pro remain competitive, they indicate potential for improvement on knowledge-intensive tasks—a reasonable trade-off given our emphasis on reasoning.

\begin{table}[H]
\begin{tcolorbox}[
    colback=tiiPurple!3!white,
    colframe=tiiPurple!75!black,
    arc=6mm,
    boxrule=0pt,
    center,       
    left=0pt,
    right=0pt,
    top=4pt,
    bottom=4pt,
    width=\textwidth,
    sharp corners=downhill
]
\centering 
\small
\color{tiiPurple!75!black}
\begin{tabular}{lcccc}
\textbf{Models} &
\textbf{GPQA-Dimond} &
\textbf{MMLU-Pro} &
\textbf{HLE} &
\textbf{IFBench}  \\
\hline\\[-1em]
Qwen3-8B & 61.2 & 63.5 & 4.2 & 35.3 \\
DeepSeek-R1-0528-Qwen3-8B & 61.4 & 69.1 & 5.6 & 29.2 \\
Phi-4-Reasoning-Plus-14B & 67.9 & 79.2 & 5.9 & 51.7 \\
Apriel-1.5-15b-Thinker & 68.2 & 76.5 & 12.0 & 55.8 \\ 
GPT-OSS-20B & 61.2 & 75.6 & {9.8} & {69.4} \\
Qwen3-32B & 67.3 & 73.9 & 8.3 & 35.4  \\
Nemotron-H-47B-Reasoning & 56.8 & 78.6 & {4.4} & 34.3 \\
\textbf{Falcon-H1R-7B} & 61.3 & 72.1 & \underline{11.1} & \underline{53.4} \\
\end{tabular}
\end{tcolorbox}
\caption{General domains reasoning. \textbf{Bold} and \underline{underline} highlight top-1 and top-2 standings of Falcon-H1R-7B w.r.t.\@ the considered baselines. Results marked with $^*$ are taken from Artificial Analysis.}
\label{tab:eval-general}
\end{table}

\begin{tcolorbox}[colback=tiiPurple!5!white, colframe=tiiPurple!75!black, title={\textbf{Competitive reasoning performance is possible with SLMs}}]
Falcon-H1R stands out for its parameter efficiency. With just 7B parameters, it consistently matches or outperforms models that are two to seven times larger—including Phi-4-Reasoning-Plus-14B, Apriel-1.5-15B-Thinker, GPT-OSS-20B, Qwen3-32B, and Nemotron-H-47B-Reasoning—on reasoning-intensive benchmarks. These results highlight how careful data curation and training strategies can deliver significant performance gains without scaling model size.
\end{tcolorbox}

\section{Test-time Scaling}\label{sec:TTS}
The parallel-hybrid design of the Falcon-H1R architecture enables efficient inference even under high batch-size settings (see Appendix \ref{sec:inference} for details). This capability is especially important for test-time scaling techniques that rely on parallel reasoning. To showcase Falcon-H1R’s performance in such scenarios, we evaluate the model using the recent Deep Think with Confidence (DeepConf) method \citep{fu2025deep}. DeepConf is an efficient test-time scaling approach that dynamically filters parallel reasoning chains based on confidence scores derived from the model itself. By terminating low-confidence reasoning paths early and allowing only high-potential chains to continue, DeepConf further reduces computational overhead.

\subsection{Setup}
The settings of our TTS evaluations are detailed as follows:
\begin{itemize}
    \item \textit{Adaptive Sampling Configuration:} We use the online algorithm \cite[Algorithm 2]{fu2025deep} with a fixed trace budget of $K=512$, where the total cost includes all generated tokens, even those from traces stopped early. The process begins with a warm-up phase that generates $N_{\text{init}}=16$ traces to determine the stopping threshold $s$, calculated by applying the Lowest Group Confidence criterion over a sliding window of 2,048 tokens. After the warm-up, we set $s$ at the $\eta$-th percentile of the minimum confidence scores from these initial traces, with $\eta=10\%$ for aggressive filtering. In the final phase, the remaining $K - N_{\text{init}} = 496$ traces are generated with early stopping: each trace is terminated once its current group confidence (computed over the most recent 2,048 tokens) falls below $s$.
    \item \textit{Generation Parameters:} All models are evaluated using their recommended decoding settings, including model-specific temperature $T$ and \text{top}-p values as suggested in their official inference guidelines. We set the maximum generation length to 64K tokens per trace, and perform inference using vLLM with tensor parallelism configured based on each model’s parameter size. Each model is conditioned with its recommended system prompt for optimal instruction-following behavior. For Phi-4-Reasoning-Plus-14B and Qwen3-8B/Qwen3-32B, we use reduced context lengths of 32K and 40K tokens, respectively, to accommodate their maximum supported window sizes.
    \item \textit{Answer Extraction:} To evaluate model output correctness, we extract final answers using the \texttt{math\_verify} parsing framework rather than relying solely on the simple \texttt{boxed\{...\}} extraction used in the original DeepConf implementation\footnote{\url{https://github.com/facebookresearch/deepconf}}. The reference DeepConf repository identifies answers by locating the last \texttt{boxed\{...\}} expression in the model output. For more robust parsing, we have consider an alternative approach: the \texttt{math\_verify} parser first tries to interpret the final mathematical expression as the ground-truth answer, enabling consistent handling of algebraic forms, numerical expressions, and non-LaTeX formatting. If parsing fails, we revert to the boxed-expression extractor.
    \item \textit{Voting Methods:} We evaluate below aggregation strategies for determining the final answer from the filtered traces:
    \begin{enumerate}
        \item \emph{Majority Voting:} selection the most frequent answer.
        \item \emph{Mean Confidence-weighted Voting:} weighting each answer by its mean group confidence.
        \item \emph{Tail Confidence-weighted Voting:} weighting answers by their minimum confidence scores.
        \item \emph{Bottom Window-weighted Voting:} uses the confidence of the lowest-scoring window.
        \item \emph{Minimum window-weighted Voting:} applies the strictest confidence-based weighting
    \end{enumerate}
    Note that across all evaluated models, we find that these voting strategies converge to very equivalent accuracies, indicating that DeepConf filtering produces a high-quality candidate trace set that is robust and flexible to the choice of the aggregation method.
\end{itemize}

\subsection{Evaluation Results}
Using the DeepConf approach with the settings describe above, we evaluate our model and competitive baselines on a selected subset of the primary mathematical and scientific reasoning benchmarks, specifically AIME-2024, AIME-2025, GPQA-D, and AMO-Bench. For AMO-Bench, we focus on the parser-verifiable subset, which comprises 39 problems with deterministic correctness verification using automated parsers. Our TTS results are summarized in the table below:

\begin{table}[H]
\centering
\footnotesize

\begin{tcolorbox}[
    colback=tiiPurple!3!white,
    colframe=tiiPurple!75!black,
    arc=6mm,
    boxrule=0pt,
    center,
    left=0pt,
    right=0pt,
    top=4pt,
    bottom=4pt,
    width=.95\textwidth,
    sharp corners=downhill
]
\centering
\color{tiiPurple!75!black}

\begin{tabular}{lcc|cc|cc|cc}
{}&
\multicolumn{2}{c}{\textbf{AIME24}} &
\multicolumn{2}{c}{\textbf{AIME25}} &
\multicolumn{2}{c}{\textbf{GPQA-D}} &
\multicolumn{2}{c}{\textbf{AMO-Bench}\footnote{\color{tiiPurple!75!black}Limited to the parser-verifiable subset which comprises 39 problems.} } \\
\textbf{Models} & \textbf{Acc.} $\uparrow$ & \textbf{Tok.} $\downarrow$ &
  \textbf{Acc.} $\uparrow$ & \textbf{Tok.} $\downarrow$ &
  \textbf{Acc.} $\uparrow$ & \textbf{Tok.} $\downarrow$ &
  \textbf{Acc.} $\uparrow$ & \textbf{Tok.} $\downarrow$ \\
\midrule

Qwen3-8B &
80.0 & 138.3 &
80.0 & 177.2 &
60.9 & 451.3 &
15.4 & 320.0 \\

DS-R1-0528-Qwen3-8B &
90.0 & 145.5 &
82.8 & 174.5 &
59.9 & 454.9 &
25.6 & 487.9 \\

Nemotron-H-8B &
53.3 & 156.0 &
43.3 & 166.8 &
61.1 & 355.0 &
7.7 & 279.4 \\

Phi-4-R-Plus-14B &
86.7 & 123.9 &
83.3 & 145.9 &
73.2 & 613.0 &
20.5 & 276.9 \\

Qwen3-32B &
86.7 & 134.4 &
86.7 & 174.8 &
70.1 & 460.0 &
28.2 & 364.8 \\

\textbf{Falcon-H1R-7B} &
\textbf{96.7} & \textbf{89.8} &
\textbf{96.7} & \textbf{95.1} &
\underline{70.2} & 452.3 &
\textbf{35.9} & \textbf{216.8} \\

\end{tabular}

\end{tcolorbox}

\caption{Test-Time Scaling Performance with DeepConf@512 \citep{deepconf} across mathematical and science benchmarks. Columns show voted accuracy (\% ± 1.5\% std) and generated tokens (M ± 5\% variation). \textbf{Bold} and \underline{underline} highlight top-1 and top-2 standings of Falcon-H1R-7B w.r.t.\@ the considered baselines.}
\label{tab:bench}
\end{table}

\noindent \textbf{Falcon-H1R} exhibits superior performance along two key dimensions: solution quality and amount of generated tokens, in addition to the computational efficiency highlighted in Appendix \ref{sec:inference}. It consistently generates
substantially fewer tokens while maintaining high accuracy across all
benchmarks, reflecting stable confidence calibration and coherent reasoning traces. These results highlight the effectiveness of our training pipeline in producing reasoning models that are both cost-efficient and highly accurate.\\

\begin{tcolorbox}[
    colback=tiiPurple!5!white,
    colframe=tiiPurple!75!black,
    title=Advancing the 3D limits of reasoning efficiency with Falcon-H1R,
    fonttitle=\bfseries
    ]
The combination of faster inference, token efficiency and higher accuracy makes Falcon-H1R-7B a practical choice for scaling advanced reasoning systems, especially in settings that require substantial parallel chain-of-thoughts generation.
\end{tcolorbox}

\section{Conclusion}
In summary, this report presented Falcon-H1R, a 7B-parameter, reasoning-optimized language model that demonstrates small language models (SLMs) can achieve state-of-the-art reasoning performance typically associated with much larger systems. Through a hybrid Transformer–Mamba architecture and a robust training strategy combining supervised fine-tuning and reinforcement learning, Falcon-H1R delivers competitive accuracy, superior inference efficiency, and substantial reductions in computational cost.
The model consistently matches or surpasses larger state-of-the-art models on a range of challenging reasoning benchmarks, validating the impact of careful data curation and targeted training via effective SFT and RL scaling. Its efficient architecture enables faster inference, greater token efficiency, and effective parallelization—features that make it well-suited for advanced reasoning tasks, especially those leveraging test-time scaling (TTS) methods.
Falcon-H1R’s integration with the DeepConf approach \citep{deepconf} further enhances its scalability and cost-effectiveness in TTS scenarios, achieving high accuracy while reducing resource consumption. These results highlight the model’s potential as a practical backbone for scalable reasoning systems in real-world applications.
Looking forward, this work opens new directions for pushing the limits of SLMs, such as training even smaller models for reasoning, and investigating architectural innovations to maximize efficiency and reliability in test-time scaling.

\bibliographystyle{apalike}   
\bibliography{referenceReasoning}

\newpage
\appendix

\section{Training Optimizations}\label{sec:train-optim}
\begin{figure}[H]
    \centering
    \includegraphics[width=.8\linewidth]{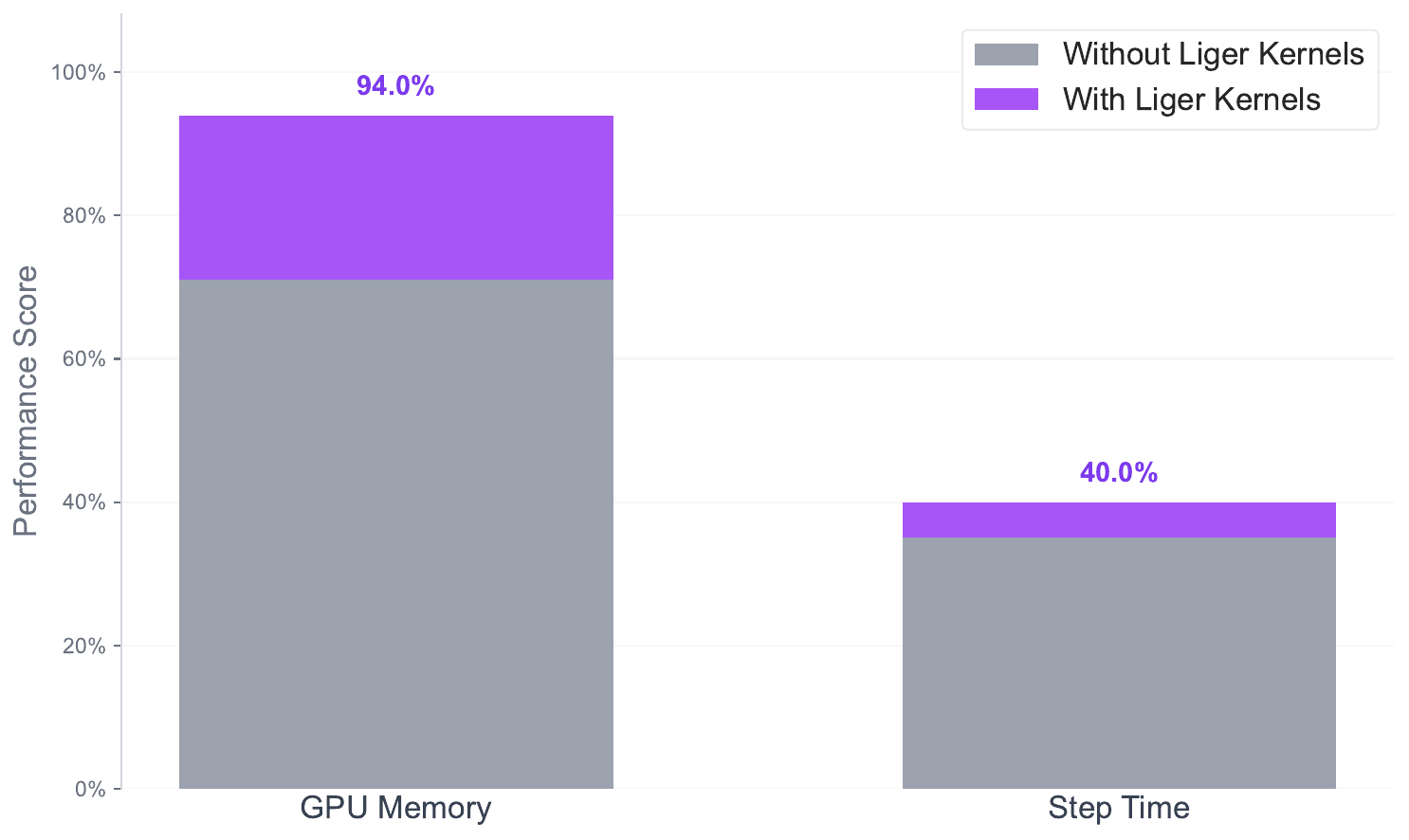}
    \caption{Comparison of GPU memory usage and per-step training time for the Falcon-H1-7B with versus without Liger Kernels.}
    \label{fig:liger_kernel_speedup}
\end{figure}

Figure~\ref{fig:liger_kernel_speedup} compares GPU memory utilization and per-step training time with and without Liger Kernels enabled. The results demonstrate consistent speedups and a reduced memory footprint for the Falcon-H1-7B model. Falcon-H1 models are now officially supported in the Liger Kernels framework~\citep{hsu2025ligerkernel} through the following upstream integrations:
\begin{itemize}
    \item \url{https://github.com/linkedin/Liger-Kernel/pull/900}
    \item \url{https://github.com/linkedin/Liger-Kernel/pull/903}
\end{itemize}

In addition to these kernel-level optimizations, we enabled fused AdamW and efficient gradient accumulation support within the \textsc{verl} framework, integrated through:
\begin{itemize}
    \item \url{https://github.com/volcengine/verl/pull/3332}
    \item \url{https://github.com/volcengine/verl/pull/3692}
\end{itemize}
Together, these improvements reduce optimizer overhead and provide smoother gradient-accumulation behavior for long-context workloads.

\section{Inference Analysis}\label{sec:inference}

\begin{table}[H]
\centering
\small
\begin{tcolorbox}[
    colback=tiiPurple!3!white,
    colframe=tiiPurple!75!black,
    arc=6mm,
    boxrule=0pt,
    center,       
    left=0pt,
    right=0pt,
    top=4pt,
    bottom=4pt,
    width=.6\textwidth,
    sharp corners=downhill
]
\centering  
\color{tiiPurple!75!black}
\begin{tabular}{lc}
\textbf{Parameter} & \textbf{Falcon-H1R-7B} \\
\hline\\[-1em]
Total Parameters & 7.59B\\
Layers & 44 \\
Hidden Dimension ($d_{\text{model}}$) & 3072 \\
Vocabulary Size & 130,048  \\
Attention Heads (Q/KV) & 12/2  \\
SSM Heads & 24 \\
Head Dimension (Attn/SSM) & 128/128  \\
State Dimension ($d_{\text{state}}$) & 256  \\
Context Length & 256K  \\
\end{tabular}
\end{tcolorbox}
\caption{Architectural specifications of Falcon-H1R-7B \citep{falconH1}.}
\label{tab:h1_arch}
\end{table}

\begin{figure}[H]
    \centering
    \includegraphics[width=.7\linewidth]{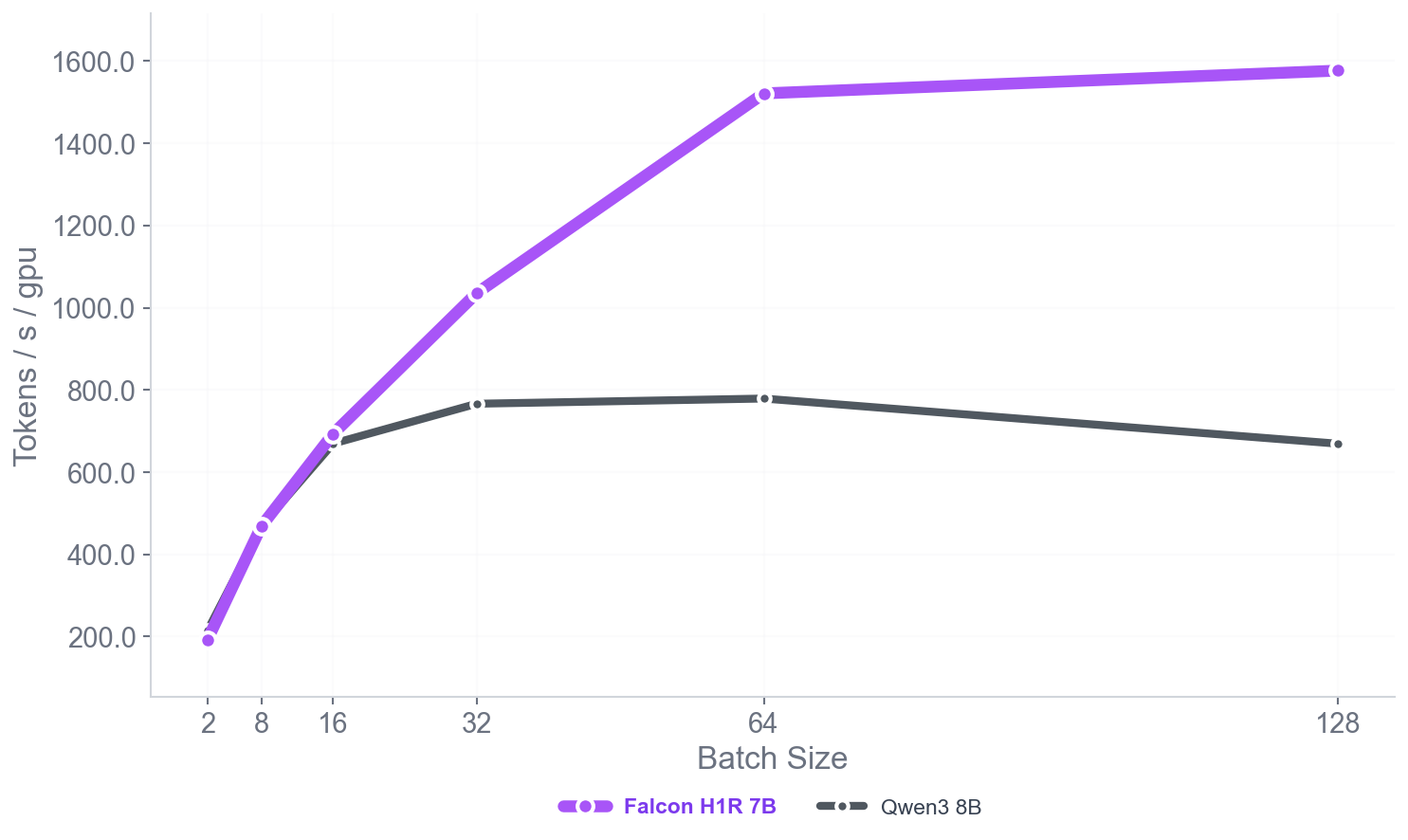}
    \caption{vLLM inference benchmarking in terms of batch size.}
    \label{fig:vllm_benchmark}
\end{figure}

Figure~\ref{fig:vllm_benchmark} presents a comparison of inference throughput using vLLM \citep{kwon2023efficient} between the transformer-based Qwen3-8B model \citep{qwen2025qwen3} and the hybrid Falcon-H1R-7B architecture (Table \ref{tab:h1_arch}). Evaluations spanned batch sizes from $2$ to $128$ and averaged two input-output token configurations: (i) $512$ input tokens with $32$K output tokens, and (ii) $8$K input tokens with $16$K output tokens. To ensure fair and representative comparisons, we selected the optimal parallelism configuration for each regime: small batches ($\text{BS}=2$) used TP1; medium batches ($\text{BS}=8$ and $\text{BS}=16$) used TP2; and the largest batches ($\text{BS}=32, 64, 128$) used TP4 for Qwen3-8B and DP2 with TP2 for Falcon-H1R-7B, maximizing throughput for each architecture. All measurements were taken on NVIDIA H100 GPUs. Falcon-H1R-7B shows clear advantages at medium to large output reasoning lengths (16K--32K) and higher batch sizes, consistently surpassing Qwen3-8B with throughput improvements of $+20\%$ to $+100\%$.

\section{RL Data Filtering Diagram}

\begin{figure}[H]
    \centering
    \includegraphics[width=\linewidth]{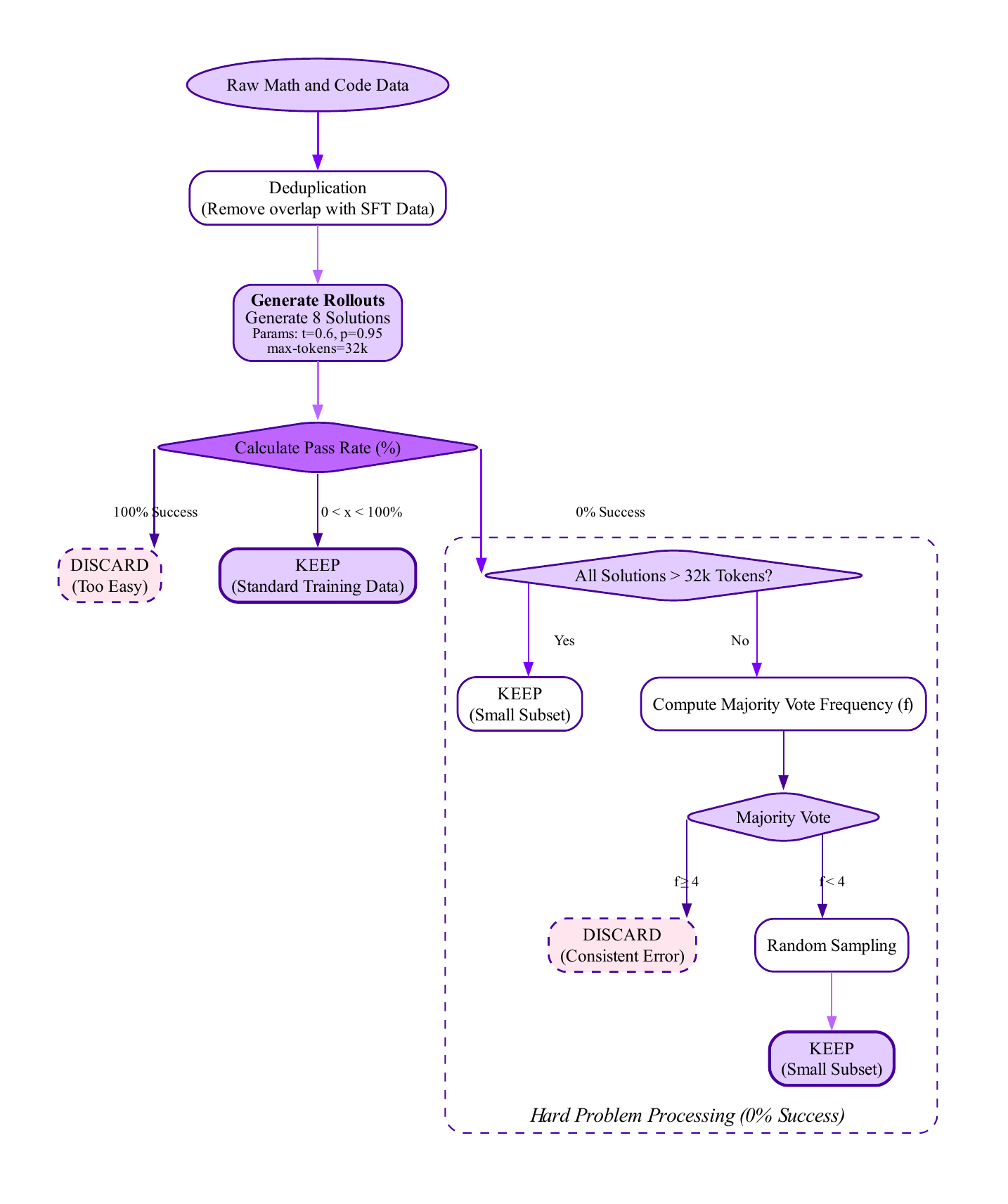}
    \caption{RL data filtering procedure as described in Section \ref{sec:data-rl}.}
    \label{fig:rl-data-diagram}
\end{figure}

\section{Benchmarks Descriptions}

\begin{table}[H]
\centering
\small
\renewcommand{\arraystretch}{1.25}
\begin{tabularx}{\textwidth}{p{2cm} p{4cm} p{8cm}}
\toprule
\textbf{Category} & \textbf{Benchmark} & \textbf{Description} \\
\midrule
\multirow{0}{2cm}{\centering Math\arraybackslash}
& AIME 2024/2025 & American Invitational Mathematics Examination problems; challenging Olympiad-style math reasoning tasks \citep{AIME2024, AIME2025}. \\
& HMMT25 & Problems from the 2025 Harvard--MIT Mathematics Tournament; recent high-school Olympiad-level problems  \citep{hmmt2025}. \\
& AMO-Bench & An Advanced Mathematical reasoning benchmark with Olympiad level or even higher difficulty, comprising 50 human-crafted problems \citep{an2025amobench}. \\
& Math500 & Curated subset of the MATH dataset with 500 medium-to-hard competition problems testing algebra, geometry, and number theory \citep{hendrycks2021measuring,lightman2023let}. \\
\midrule
\multirow{0}{2cm}{\centering Code\arraybackslash}
& LiveCodeBench & Python programming to solve programming scenarios derived from LeetCode, AtCoder, and Codeforces \citep{LCB}. \\
& SciCode & Tests the model’s ability to reason scientifically and generate executable scientific code  \citep{scicode}. \\
& Terminal-Bench Hard  & Measures multi-step planning and tool use in realistic command-line environments \citep{tbench_2025}. \\
& $\tau^2$-Bench Telecom & Domain-specific benchmark assessing agentic reasoning and task execution in telecom scenarios \citep{barres2025tau2}. \\
\midrule
\multirow{0}{2cm}{\centering General\arraybackslash}
& GPQA-Diamond & Graduate-level question set across physics, chemistry, and biology requiring multi-step scientific reasoning \citep{gpqa}. \\
& Humanity’s Last Exam & Frontier-level benchmark of extremely challenging reasoning questions across STEM and humanities \citep{humanitysexam}. \\
& MMLU-Pro & Advanced version of MMLU covering 14 domains with higher difficulty and reasoning depth \citep{wang2024mmlupro}. \\
& IFBench & Tests precise instruction-following, constraint satisfaction, and output format compliance \citep{pyatkin2025ifbench}. \\
\bottomrule
\end{tabularx}
\caption{Overview of the considered evaluation benchmarks.}
\label{tab:benchmarks}
\end{table}

\section{Safety Evaluation}\label{sec:safety}
We assess the safety performance of the Falcon-H1R-7B model using a comprehensive set of benchmarks that cover jailbreak attempts, adversarial prompts, and harmful content detection. To gain a thorough understanding of model safety across different output configurations, we use three evaluation modes: (1) \textbf{CoT Only}, which evaluates only the chain-of-thought reasoning for safety violations while ignoring the final answer; (2) \textbf{Answer Only}, which assesses only the final answer, excluding the reasoning process; and (3) \textbf{CoT+Answer}, which jointly evaluates both the reasoning trace and the final answer for safety. This multi-pronged strategy reveals whether safety issues arise mainly during the model's reasoning, its final outputs, or both, offering insights into where interventions are most effective. As summarized in Table~\ref{tab:safety_eval7BRL}, we tested the Falcon-H1R-7B model on $81,970$ prompts drawn from diverse safety benchmarks, including JailbreakBench \citep{jailbreakbench}, ALERT \citep{alert}, AllenAI \citep{wildguard}, WalledEval \citep{walledeval}, and SALAD Bench \citep{salad}. Safety scores were assigned using the \texttt{meta-llama/Llama-Guard-3-8B} safety classifier \citep{llama3}. 

\vspace{.7cm}


\begin{table}[H]
\centering

\begin{tcolorbox}[
    colback=tiiPurple!3!white,
    colframe=tiiPurple!75!black,
    arc=6mm,
    boxrule=0pt,
    center,       
    left=0pt,
    right=0pt,
    top=4pt,
    bottom=4pt,
    width=\textwidth,
    sharp corners=downhill
]
\centering  
\small
\color{tiiPurple!75!black}
\begin{tabular}{lcccc}
\textbf{Benchmark} & \textbf{\# Prompts} & \textbf{CoT Only} & \textbf{CoT+Answer} & \textbf{Answer Only} \\
\midrule
\multicolumn{5}{l}{\textbf{JailbreakBench} \citep{jailbreakbench}} \\
\hspace{0.5em} Behaviors  & 100 & 94\% & 96.0\% & 97.0\% \\
\hspace{0.5em} Judge comparison  & 300 & 96.67\% & 98.67\% & 98.67\% \\
\hspace{0.5em} Jailbreak prompts  & 50 & 80\% & 90.0\% & 90.0\% \\
\midrule
\multicolumn{5}{l}{\textbf{ALERT} \citep{alert}} \\
\hspace{0.5em} Non-adversarial  & 14763 & 90.92\% & 97.7\% & 97.72\% \\
\hspace{0.5em} Adversarial  & 30966 & 92.44\% & 98.15\% & 98.19\% \\
\midrule
\multicolumn{5}{l}{\textbf{AllenAI} \citep{WildJailbreak}}  \\
\hspace{0.5em} Wildjailbreak  & 2210 & 96.1\% & 98.51\% & 98.31\% \\
\midrule
\multicolumn{5}{l}{\textbf{WalledEval} \citep{walledeval}}  \\
\hspace{0.5em} AdvBench  & 520 & 97.67\% & 99.42\% & 99.42\% \\
\hspace{0.5em} AyaRedTeaming  & 987 & 89.06\% & 96.15\% & 96.15\% \\
\hspace{0.5em} BeaverTails  & 700 & 89.68\% & 97.14\% & 97.14\% \\
\hspace{0.5em} CatHarmfulQA  & 550 & 82.16\% & 97.82\% & 98.0\% \\
\hspace{0.5em} HarmBench contextual  & 100 & 94\% & 98.0\% & 98.0\% \\
\hspace{0.5em} HarmBench copyright  & 100 & 100.0\% & 100.0\% & 100.0\% \\
\hspace{0.5em} HarmBench standard & 200 & 92\% & 99.0\% & 99.0\% \\
\hspace{0.5em} Stereotype & 3456 & 93.89\% & 98.96\% & 98.96\% \\
\hspace{0.5em} XSTest & 450 & 87.07\% & 96.44\% & 98.0\% \\
\midrule
\multicolumn{5}{l}{\textbf{Salad Bench} \citep{salad}} \\
\hspace{0.5em} Attack enhanced set & 5000 & 95.13\% & 97.86\% & 97.86\% \\
\hspace{0.5em} Defense enhanced set & 200 & 97\% & 99.5\% & 99.5\% \\
\hspace{0.5em} Base set & 21318 & 93.29\% & 98.6\% & 98.58\% \\
\midrule
\textbf{SIMPLE AVERAGE} & \textbf{81970} & \textbf{92.23\%} & \textbf{97.66\%} & \textbf{97.71\%} \\
\textbf{WEIGHTED AVERAGE} & \textbf{81970} & \textbf{92.60\%} & \textbf{98.18\%} & \textbf{98.19\%} \\
\end{tabular}
\end{tcolorbox}

\caption{
Safety evaluation results of the Falcon-H1R-7B model across multiple benchmarks.}
\label{tab:safety_eval7BRL}
\end{table}

\paragraph{Summary of Safety Evaluation:} We conducted a comprehensive safety evaluation of the Falcon-H1R-7B model across 81,970 prompts, yielding important insights into the safety characteristics of reasoning models. The Falcon-H1R-7B model exhibits strong safety performance, with weighted averages of 98.18\% for CoT+Answer and 98.19\% for Answer Only, indicating robust safety alignment in its final outputs.

\paragraph{Reasoning vs. Final Answers:} A key observation is the anticipated safety gap between chain-of-thought (CoT) reasoning (92.60\% weighted average) and final answers (98.19\%). This 5.59 percentage point difference arises naturally from the reasoning process. During the CoT phase, the model explores multiple perspectives, evaluates different approaches, and reasons through the problem space, including understanding why certain requests might be harmful or how they could be misused. This exploratory process often requires engaging with sensitive content at a deeper level than simply producing a final response. For example, when faced with a potentially harmful query, the model may reason about the nature of the harm, consider edge cases, or evaluate the intent behind the request before arriving at an appropriate refusal. As a result, safety classifiers are more frequently triggered during reasoning than when evaluating only the final answer. This pattern is observed across nearly all benchmarks.

\paragraph{Benchmark Case Studies:}
\begin{itemize}
    \item \textit{CatHarmfulQA}: CoT safety is 82.16\%, while Answer safety reaches 98.0\%. The model reasons extensively about harmful content categories before issuing a safe refusal.
    \item \textit{XSTest}: 87.07\% CoT vs. 98.0\% Answer. The model analyzes potentially harmful scenarios before ultimately declining assistance.
    \item \textit{BeaverTails}: 89.68\% CoT vs. 97.14\% Answer. The reasoning process involves engagement with unsafe content, but the final output is safe.
\end{itemize}

Lower CoT safety scores do not indicate a safety failure. Instead, they reflect the model's thorough process of engaging with and understanding unsafe content in order to provide appropriate refusals. The consistently high Answer Only scores across all benchmarks (90.0\% to 100.0\%) demonstrate that the model effectively refuses or safely responds to harmful queries in its response.

\paragraph{Implications for Deployment:} The separation between reasoning and response functions as intended: the model carefully reasons—sometimes about unsafe content—before producing a safe, appropriate answer. For deployment, this distinction suggests that exposing raw reasoning traces to end users requires careful consideration. While reasoning traces are valuable for transparency and interpretability, their exploratory nature means they may contain content that, although part of legitimate safety reasoning, could appear concerning out of context. Organizations may prefer to show only final answers to users while retaining reasoning traces for auditing, debugging, or research purposes.

\begin{tcolorbox}[colback=tiiPurple!5!white, colframe=tiiPurple!75!black, title={\textbf{CoT-Answer safety gap reflects deliberative reasoning, not safety failure}}]
The safety gap between chain-of-thought reasoning and final answers is a natural consequence of the model's deliberative process. During CoT, the model engages with and analyzes potentially harmful content to understand why requests are problematic—this exploratory reasoning triggers safety classifiers more frequently, while final outputs consistently demonstrate robust safety alignment.
\end{tcolorbox}

\end{document}